\documentclass{article}

\usepackage{microtype}
\usepackage{graphicx}
\usepackage{subcaption}
\usepackage{booktabs} 

\usepackage{hyperref}

\usepackage[accepted]{icml2026}

\usepackage[table]{xcolor}
\usepackage{enumitem}
\usepackage{array}
\usepackage{mathrsfs}
\usepackage{amsthm}
\usepackage{amsmath}
\usepackage{amsfonts}
\usepackage{multirow}
\usepackage{color}
\usepackage{arydshln}
\usepackage{makecell}
\usepackage{bbding}
\usepackage{colortbl}
\usepackage{amssymb}
\usepackage{mathtools}
\usepackage{tcolorbox}  

\usepackage[capitalize,noabbrev]{cleveref}

\theoremstyle{plain}

\theoremstyle{definition}

\theoremstyle{remark}

\usepackage[textsize=tiny]{todonotes}

\icmltitlerunning{Activation Steering Meets Reinforcement Unlearning for Multimodal Large Language Models}

\begin{document}

\twocolumn[
 \icmltitle{ASRU: Activation Steering Meets Reinforcement Unlearning for Multimodal Large Language Models}


  \icmlsetsymbol{equal}{*}

  \begin{icmlauthorlist}
    \icmlauthor{Jiahui Guang}{hit,pcl}
    \icmlauthor{Haiyan Wang}{pcl}
    \icmlauthor{Yingjie Zhu}{hit,pcl}
    \icmlauthor{Cuiyun Gao}{hit}
    \icmlauthor{Jing Li}{polyu,polyu-huizhou}
    \icmlauthor{Di Shao}{pcl}
    \icmlauthor{Zhaoquan Gu}{hit,pcl}

  \end{icmlauthorlist}

  \icmlaffiliation{hit}{Harbin Institute of Technology, Shenzhen, China}
  \icmlaffiliation{pcl}{Peng Cheng Laboratory, Shenzhen, China}
  \icmlaffiliation{polyu}{The Hong Kong Polytechnic University, Hong Kong, China}
  \icmlaffiliation{polyu-huizhou}{The Hong Kong Polytechnic University-Daya Bay Technology and Innovation Research Institute, Huizhou, China}

  \icmlcorrespondingauthor{Cuiyun Gao}{gaocuiyun@hit.edu.cn}

  \icmlkeywords{Machine Learning, ICML}

  \vskip 0.3in
]



\printAffiliationsAndNotice{}  

\begin{abstract}
Multimodal large language models (MLLMs) may memorize sensitive cross-modal information during pretraining, making machine unlearning (MU) crucial. Existing methods typically evaluate unlearning effectiveness based on output deviations, while overlooking the generation quality after unlearning. This can easily lead to hallucinated or rigid responses, thereby affecting the usability and safety of the unlearned model. To address this issue, we propose \textbf{ASRU}, a controllable multimodal unlearning framework that incorporates generation quality as a core evaluation objective. ASRU first induces initial refusal behavior through activation redirection, and then optimizes fine-grained refusal boundaries using a customized reward function, thereby achieving a better trade-off between target knowledge unlearning and model utility. Experiments on Qwen3-VL show that ASRU significantly improves unlearning effectiveness (+24.6\%) on average and generation quality (5.8×) on average  while effectively preserving model utility, using only a small amount of retained supervision data. The code for ASRU is released at \url{https://github.com/guangjh/ASRU}.

\end{abstract}

\section{Introduction}\label{sec:1}
Multimodal Large Language Models (MLLMs) have achieved remarkable success across a wide range of multimodal tasks \citep{huang2023visual,wang2024qwen2,zou2025deep,du2025makes,zhu-etal-2025-benchmarking,hu2025praxis}.
However, their large-scale pretraining often relying on millions of instances—inevitably causes MLLMs to memorize sensitive or harmful information \citep{karamolegkou-etal-2023-copyright,huang-etal-2024-demystifying}, raising serious concerns related to privacy leakage and copyright infringement \cite{huo-etal-2025-mmunlearner,liu2025protecting}.
Since pretraining data are typically inaccessible, retraining models from scratch to remove such knowledge is impractical, making post-hoc machine unlearning a critical requirement for building trustworthy MLLM systems \citep{liu2025rethinking,chen2025safeeraser,ding2025mllmeraser}.

\begin{figure}
    \centering
    \includegraphics[width=\linewidth]{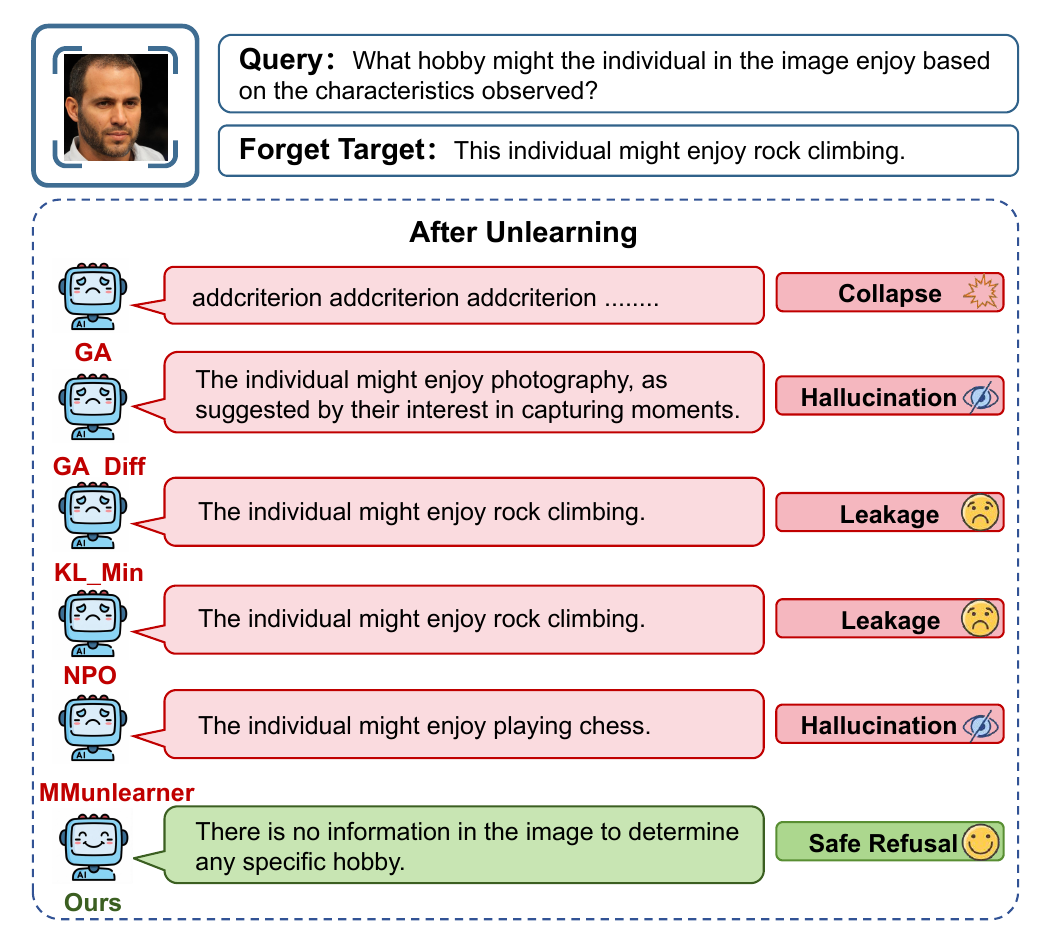}
    \caption{Existing unlearning methods often yield \textbf{hallucinated}, \textbf{rigid outputs} and  \textbf{exhibit incomplete unlearning}, whereas \textbf{our ASRU} produces context-aware dynamic refusals.}
    \label{fig:1}
\end{figure}

Most existing work directly adapts unlearning strategies designed for text-only LLMs to the multimodal setting, including gradient-ascent-based methods \citep{thudi2022unrolling,liu2024towards,zhang2024negative,li2024single}, preference optimization \cite{zhang2024negative}, targeted parameter updates \citep{huo-etal-2025-mmunlearner,li-etal-2025-forget}.
Despite their initial progress, these methods still suffer from two fundamental limitations. As shown in Figure~\ref{fig:1}, \textbf{(I) Unnatural and hallucinated responses after unlearning remain prevalent.} Existing methods either generate fabricated content that is inconsistent with the target facts, or produce abnormal, repetitive, and incoherent degenerated text. In high-risk domains, hallucinated outputs may introduce severe safety hazards, such as generating plausible but incorrect medical advice after patient records have been removed \cite{shen2025llm}. Meanwhile, unnatural responses may inadvertently expose that an unlearning operation has been performed, which could introduce potential security risks. We refer to this phenomenon as \textbf{post-unlearning responsiveness degradation:} after losing the target knowledge, the model fails to express its knowledge gap in a natural, context-aware, and coherent manner. We argue that an unlearned model should behave like an aligned base model, accurately expressing its knowledge gaps in a dynamic, context-aware, and coherent manner. Accordingly, we advocate treating \textbf{generation quality} as a primary evaluation criterion in unlearning research. \textbf{(II) A persistent trade-off exists between unlearning quality and model utility.}
Gradient ascent based update methods tend to over-suppress the forget set 
\(D_f\), leading to unintended global behavior shifts and utility degradation \cite{wang2025rethinking}.
Conversely, methods that preserve utility often exhibit incomplete unlearning, making it difficult to achieve an optimal trade-off between effective unlearning and utility preservation.

These limitations indicate that existing methods still lack fine-grained control over the refusal boundary and have not sufficiently addressed generation quality after unlearning. To address these challenges, we propose Activation Steering meets Reinforcement Unlearning (ASRU), a novel framework for multimodal unlearning. ASRU first constructs a knowledge-absence direction and performs activation steering, inducing a basic refusal behavior by updating only a single down-projection matrix. We further introduce Group Relative Policy Optimization (GRPO) based on this initialization, design a verifiable reward function, and leverage a small supervised retain subset to jointly optimize forget samples with their highly related boundary samples, thereby learning a context-aware, fine-grained refusal boundary. Experimental results show that ASRU achieves fine-grained boundary perception, generates contextually dynamic and consistent refusal responses, and obtains a better trade-off between unlearning effectiveness and model utility.

We highlight our main contributions as:
\begin{itemize}
\item Existing multimodal unlearning methods assess unlearning primarily via deviation from ground truth. We introduce \textbf{generation quality} as a key evaluation dimension, capturing whether unlearned model produce coherent, context-aware, and natural refusals.
\item We propose \textbf{ASRU}, a novel and controllable unlearning framework that first integrates activation steering with GRPO. ASRU first uses activation steering (via updating a single projection matrix) to induce controllable refusal behavior as a initialization, and then introduces a boundary set and verifiable reward design to learn a finer-grained forget–retain boundary.
\item Experimental results demonstrate that ASRU preserves model utility while achieving up to an average 5.8× improvement in generation quality and an average 24.61\% gain in unlearning effectiveness over existing baselines, with a superior forget--retain trade-off using only a small amount of supervised retain data.
\end{itemize}

\section{Related work}
\subsection{Multimodal Unlearning}
Machine unlearning in multimodal large language models remains relatively underexplored \cite{li2025llmunlearningllmbeliefs}.
SIU \cite{li2024single} is the first work to study multimodal unlearning, focusing on removing visual recognition capabilities for specific concepts, but it relies on complex and multifaceted fine-tuning data.
MLLMU-Bench \cite{liu2025protecting} introduces a benchmark based on fictitious personal profiles and systematically evaluates multiple unlearning strategies.
MMUnlearner \cite{huo-etal-2025-mmunlearner} selectively updates specific model parameters to induce forgetting, while other approaches apply activation steering solely at inference time \cite{ding2025mllmeraser}.

\subsection{Activation Steering}
Activation (or representation) steering is a core paradigm in representation engineering \citep{yunfan2025mitigating,sterz2025recover,stolfo2024improving,stoehr-etal-2024-activation,zhanguncovering}, aiming to identify concept-specific directions (e.g., truthfulness or toxicity) in model representations and intervene on hidden states accordingly.
Its effectiveness is often explained by the linear representation hypothesis, which assumes that hidden representations can be approximated as linear combinations of attribute vectors \cite{park2023linear}.
In practice, concept directions are obtained via contrastive activation steering \citep{jorgensen2023improving,rimsky-etal-2024-steering,stoehr-etal-2024-activation} or linear probing \citep{jorgensen2023improving,yunfan2025mitigating,sterz2025recover}, and applied at inference time by modifying hidden states \citep{singh2024representation,ding2025mllmeraser,sheng2025alphasteer}.

\subsection{Reinforcement Learning for LLM Alignment}
Reinforcement learning has become a powerful post-training technique for aligning LLM behavior \citep{dai2023safe,wu2025activation,niu2025mitigatingsafetyalignmenttax}.
PPO \cite{schulman2017proximalpolicyoptimizationalgorithms} formulates generation as an MDP \cite{bellman1957markovian} but is computationally expensive.
DPO \cite{rafailov2023direct} and GRPO \cite{shao2024deepseekmath} reduce reliance on explicit reward models by leveraging preference comparisons or group-relative advantages.
Recent work explores applying RL to representation-level alignment \cite{wu2025activation} and safety optimization with constrained policies \cite{niu2025mitigatingsafetyalignmenttax}.
ASRU builds on these advances by using GRPO to learn a verifiable and fine-grained unlearning boundary in multimodal unlearning.

\begin{figure*}[t]
    \centering
        \includegraphics[width=\linewidth]{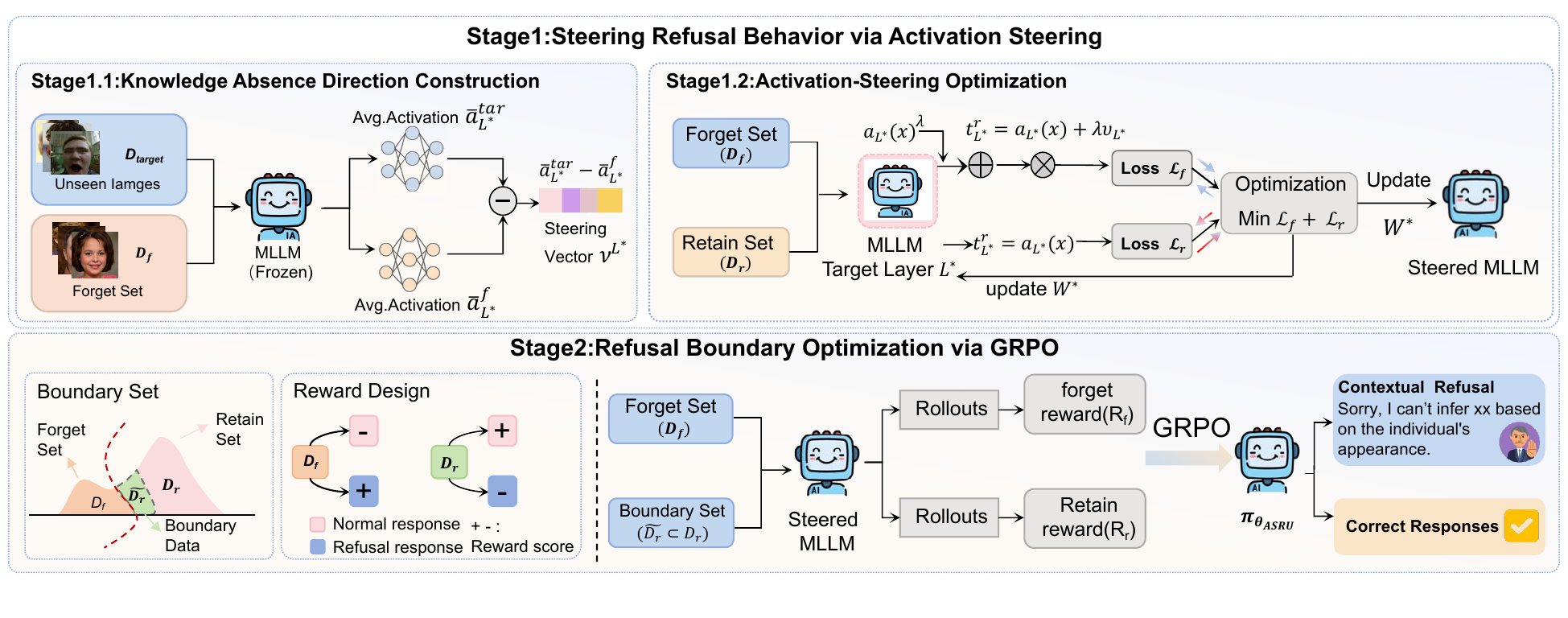}
    \caption{Overview of the proposed ASRU framework. \textbf{Stage 1} induces refusal capability via activation steering and consists of two sub-stages: \textbf{Stage 1.1} identifies the direction of knowledge absence by contrasting the model's activation patterns on previously unseen images against those on the forget set; \textbf{Stage 1.2} steers the model by training only the down-projection module at layer $L^{*}$, thereby promoting the emergence of a refusal-style response behavior. \textbf{Stage 2} then applies GRPO optimization on the boundary set and the forget set, enabling the model to learn a finer-grained refusal decision boundary.}
    \label{fig:overview ASRU}
\end{figure*}

\section{Method}
\subsection{Problem Setup}\label{sec:3.1}
Given a vision language instruction dataset
\(\mathcal{D}=\{(I_i,Q_i,A_i)\}_{i=1}^{N}\), where \(I_i\) denotes the input image, \(Q_i\) is the corresponding textual instruction, and
\(A_i=(y_1,y_2,\dots,y_{T_i})\) is the target output sequence conditioned on \((I_i,Q_i)\), it maximizes the conditional likelihood of each token \(y_t\):
\vspace{-4pt}
\begin{equation}
\max_{\theta}\ \sum_{t=1}^{T_i}\log p_{\theta}\!\left(y_t \mid y_{<t}, I_i, Q_i\right).
\end{equation} 

The goal of multimodal unlearning is to selectively remove memory associated with specific target knowledge from a pretrained model $M_{\mathrm{org}}$. The unlearning process aims to obtain an updated model $M_{\mathrm{unlearn}}$ that no longer retains the target knowledge while preserving the original multimodal understanding capability and overall utility.  

\begin{tcolorbox}[title=MLLM Unlearning Definition]
MLLM unlearning is defined as the process of modifying a MLLM to forget image-paired knowledge associated with the forget set $\mathcal{D}_f$ while preserving model utility.
\end{tcolorbox}

For the original multimodal large model \(M\), the goal of an unlearning algorithm is to obtain an updated model \(M'\). Ideally, \(M'\) should behave similarly to a model trained only on the retain set \(\mathcal{D}_r\) and never exposed to the forget set \(\mathcal{D}_f\). The model should forget the target knowledge in \(\mathcal{D}_f\) while preserving model utility on \(\mathcal{D}_r\). This process can be formally expressed as follows:

\begin{equation}
\begin{split}
\min_{\theta} \;
& \mathbb{E}_{(I_f, Q_f, A_f) \sim \mathcal{D}_f} \big[ \ell_f(A_f \mid I_f, Q_f; \theta) \big] \; + \\
& \lambda \, \mathbb{E}_{(I_r, Q_r, A_r) \sim \mathcal{D}_r} \big[ \ell_r(A_r \mid I_r, Q_r; \theta) \big],
\end{split}
\end{equation}

where \(\ell_f\) denotes the forgetting loss that discourages the model from producing target knowledge, \(\ell_r\) denotes the retention loss that preserves general multimodal performance, and \(\lambda\) controls the trade-off between forgetting and retention.

\subsection{ASRU}
As discussed in Sections~\ref{sec:1} and~\ref{sec:3.1}, effective MLLM unlearning should enable a \emph{refuse-when-necessary} and \emph{answer-when- appropriate} behavior. Specifically, the model should produce context-aware refusals for forget-set inputs while still answering permissible queries. This requires learning a precise refusal boundary between forget-related inputs and retained inputs.
To address the challenges, we propose \textbf{ASRU}, a two-stage framework that equips MLLMs with controllable refusal behavior while preserving overall utility. As illustrated in Figure~\ref{fig:overview ASRU}, in the first stage, ASRU uses activation steering and replaces the standard cross-entropy loss with a local representation loss on residual-stream activations. By updating only a single projection matrix, it induces controllable initial refusal behavior and obtains the initialized steered model \(M_{\mathrm{steered}}\).
To enable the model to learn a finer-grained refusal boundary, the second stage introduces a boundary set and a verifiable reward design to further optimize the fine-grained forget--retain boundary between forget samples and retain samples.
The core idea of ASRU is to first establish a refusal prototype and then refine the decision boundary.
The algorithm pseudo-code is provided in Appendix~\ref{algorithm}.


\subsubsection{Refusal Steering}\label{3.2.1}

A core challenge in controllable multimodal unlearning is that pretrained MLLMs typically do not spontaneously exhibit refusal behavior. To equip the model with this capability, we adopt activation steering to redirect the model’s internal representations associated with the forget set toward a refusal direction state, thereby inducing initial refusal behavior.

\textbf{Refusal Direction Construction.}
The ideal behavior of an unlearned model should be consistent with its knowledge boundary: for queries related to the forget set, the model should behave like a retrained model that has never learned the corresponding knowledge, and clearly and naturally express that \textit{it cannot provide the relevant information}. To achieve this goal, we redirect the activations of forget-set samples at the last token position toward a reference direction for refusal responses, guiding the model on the forget set toward a representational region of ``no relevant knowledge and refusal.''
Specifically, we estimate the refusal reference direction using the mean hidden activations of two contrastive groups. For a designated layer \(L^*\), we compute the average activations of the forget set \(\mathcal{D}_f\) and the target refusal-reference set \(\mathcal{D}_{\text{target}}\), respectively. Here, \(\mathcal{D}_{\text{target}}\) consists of inputs for which the model has no prior knowledge, such as images of individuals not seen during pretraining paired with privacy-related queries\footnote{Here, we employ the synthetic face images from DigiFace1M~\citep{bae2023digiface1m} as the unseen images in \(\mathcal{D}_{\text{target}}\).}. Mathematically, the mean activations are expressed as:

\vspace{-4pt}
\begin{equation}
\begin{cases}
\bar{a}_{L^*}^{\text{target}} = \frac{1}{|\mathcal{D}_{\text{target}}|} \sum_{x \in \mathcal{D}_{\text{target}}} h^{(l)}(x),\\[1ex]
\bar{a}_{L^*}^{f} = \frac{1}{|\mathcal{D}_f|} \sum_{x \in \mathcal{D}_f} h^{(l)}(x),
\end{cases}
\end{equation}

\textit{steering vector} is defined by the difference of these means:
\vspace{-4pt}
\begin{equation}
v_{L^*} = \bar{a}_{L^*}^{\text{target}} - \bar{a}_{L^*}^{f}.
\end{equation}

\textbf{Activation Steering Optimization.}  
We replace the standard cross-entropy loss with a local representation loss on residual-stream activations to perform activation redirection. Specifically, for forget-set inputs, we steer the activations along the refusal reference direction, while for retain-set inputs, we preserve their original activations:
\vspace{-4pt}
\begin{equation}
t_{L^*}^{f} = a_{L^*}(x) + \lambda v_{L^*}, \quad
t_{L^*}^{r} = a_{L^*}(x),
\end{equation}

where $\lambda$ modulates the influence of the steering vector. The model is then optimized to minimize the deviation between its actual activations and the target activations for both forget and retain sets:
\vspace{-4pt}
\begin{align}
\mathcal{L}_f &= \mathbb{E}_{x \in \mathcal{D}_f} \| a_{L^*}^f(x) - t_{L^*}^{f}(x) \|_2^2, \\
\mathcal{L}_r &= \mathbb{E}_{x \in \mathcal{D}_r} \| a_{L^*}^r(x) - t_{L^*}^{r}(x) \|_2^2, \\
\mathcal{L} &= \mathcal{L}_f + \mathcal{L}_r.
\end{align}

\textbf{Closed-form Solution via Down-projection.}  
This activation redirection can be efficiently achieved by updating only the \textbf{down-projection matrix} of a single MLP layer, while keeping the majority of the base model parameters frozen. Let \(H_f\) and \(H_r\) denote the inputs to this down-projection layer for the forget and retain sets, respectively, and let \(T_f\) and \(T_r\) denote the corresponding redirected target activations.
The optimization objective is defined as:
\vspace{-4pt}
\begin{equation}
    W^{*} = \arg\min_{W}\left\| [H_f,\, H_r]\, W - [T_f,\, T_r] \right\|_{2}.
\end{equation}
\noindent where $W^{*}$ denotes the weights of the down-projection layer.
It has been proven that this optimization problem admits a closed-form solution~~\citep{shen2025llm}:
\vspace{-4pt}
\begin{align}
\begin{split}
W^* = \left([H_f, H_r]^\top [H_f, H_r] + \gamma I\right)^{-1} 
[H_f, H_r]^\top [T_f, T_r].
\end{split}
\end{align}
where $\gamma \ge 0$ is the regularization parameter.

\subsubsection{Refusal Boundary Optimization}\label{3.2.2}

However, although the steered model \(M_{\mathrm{steered}}\) has acquired refusal capability, it still struggles to clearly distinguish the forgetting boundary, which leads to unnecessary refusals on the retain set.
To enable the model to learn a finer-grained forgetting boundary and thus dynamically and accurately exhibit a ``refuse-when-necessary, answer-when-appropriate'' response pattern, we propose a reinforcement-learning-based \emph{Forget Boundary Optimization Module}.
By introducing a boundary set and a verifiable reward design, this module further optimizes the model behavior, encouraging the model to effectively forget target knowledge while preserving its normal answering ability for non-target knowledge as much as possible.

\textbf{Boundary Set Construction.}
A key requirement in MLLM unlearning is to accurately delineate the decision boundary between forget samples and retain samples. To this end, we sample a \textit{boundary set} \(\tilde{\mathcal{D}}_r \subset \mathcal{D}_r\) from the retain set, whose samples are highly similar to \(\mathcal{D}_f\) in the input space but should be retained, enabling the model to learn a fine-grained refusal boundary.

\textbf{Reward Definition.}  
Let $(x_f, x_{\tilde{r}})$ denote paired inputs from the forget set and boundary set, with corresponding outputs $(y_f, y_{\tilde{r}})$. We define the reward function as:
\vspace{-4pt}
\begin{equation}
R(x_f, x_{\tilde{r}}, y_f, y_{\tilde{r}}) = r_f(x_f, y_f) + r_{\tilde{\mathcal{D}}_r}(x_{\tilde{r}}, y_{\tilde{r}}),
\end{equation}

where $r_f$ quantifies the effectiveness of forgetting on $\mathcal{D}_f$, and $r_{\tilde{\mathcal{D}}_r}$ evaluates retention fidelity on the boundary set. 

To stabilize policy gradient updates, we define the advantage function based on within-group relative rewards:
\vspace{-4pt}
\begin{equation}
A_i = R_i - \frac{1}{G}\sum_{j=1}^{G} R_j, \quad 
R_i = R(x, y_f^{(i)}, y_{\tilde{r}}^{(i)}),
\end{equation}

\begin{table*}[t]
\centering
\caption{Unlearning performance on MLLMU-Bench (5\% forget). $\tilde{D_r}$ represents the number of samples required from the supervised retain set.
\textcolor{red}{$\boldsymbol{\downarrow}$} indicates that lower values are preferred, while \textcolor{red}{$\boldsymbol{\uparrow}$} indicates that higher values are preferred.}
\resizebox{\linewidth}{!}{
\renewcommand\arraystretch{1.1}
\begin{tabular}{lccccccccccc}
\toprule
\multirow{2}{*}{\textbf{Method}} & \multirow{2}{*}{\textbf{Samples}}  & \multicolumn{4}{c}{\textbf{Forget Quality} \textcolor{red}{$\boldsymbol{\downarrow}$ }}                                                         & \multicolumn{2}{c}{\multirow{2}{*}{\textbf{Generation Quality} \textcolor{red}{$\boldsymbol{\uparrow}$ }}} & \multicolumn{4}{c}{\textbf{Retain Quality} \textcolor{red}{$\boldsymbol{\uparrow}$ }}                                                                   \\ \cmidrule(lr){3-6} \cmidrule(lr){9-12}
\textbf{}            & \textbf{}         & \multicolumn{2}{c}{\textbf{Class.}} & \multicolumn{2}{c}{\textbf{Gen.}} & \multicolumn{2}{c}{}                                             & \multicolumn{2}{c}{\textbf{Class.}} & \multicolumn{2}{c}{\textbf{Gen.}} \\ \midrule
\textbf{}            & \textbf{$\tilde{D_r}$} & \textbf{Forget}              & \textbf{Test}             & \textbf{Forget}          & \textbf{Test}           & \textbf{CR}       & \textbf{Forgetfulness}       & \textbf{Retain}             & \textbf{Real}              & \textbf{Retain}          & \textbf{Real}           \\\midrule

\rowcolor{gray!10} \multicolumn{12}{c}{\textbf{Qwen3-VL-4B(5\%Forget)-VQA}}         \\ 

\textcolor{gray}{\textbf{Vanilla}}     &\textcolor{gray}{-}                 & \textcolor{gray}{46.67}                       & \textcolor{gray}{50.00}                      & \textcolor{gray}{0.578}                    &\textcolor{gray}{ 0.327}                   &\textcolor{gray}{ -}                                 & \textcolor{gray}{-}                            & \textcolor{gray}{41.95 }                      &\textcolor{gray}{ 63.58 }                     &\textcolor{gray}{ 0.541 }                   &\textcolor{gray}{ 0.428 }                  \\
\textbf{GA}          & 0\%               & 40.83                       & 45.00                      & 0.511                    & 0.288                   & 0.12                              & 0.14                         & 39.41                       & 62.14                      & 0.517                    & 0.377                   \\
\textbf{GA\_diff}    & 100\%             & 45.83                       & 50.00                      & 0.548                    & 0.312                   & 0.06                              & 0.10                         & 34.50                       & 56.40                      & 0.512                    & 0.370                   \\
\textbf{NPO}         & 0\%               & 40.00                       & 43.33                      & 0.561                    & 0.327                   & 0.12                              & 0.22                         & 32.56                       & 58.75                      & 0.520                    & 0.371                   \\
\textbf{KL\_Min}     & 100\%             & 44.80                       & 46.67                      & 0.549                    & 0.315                   & 0.26                              & 0.30                         & 38.48                       & 61.36                      & 0.518                    & 0.374                   \\
\textbf{MMunlearner} & 100\%          & 40.00                       & 47.20                      & 0.552                    & 0.298                   & 0.01                              & 0.02                         & 39.28                       & 62.27                      & 0.527                    &0.380                    \\
\rowcolor{blue!10} \textbf{ASRU(ours)}  & 27.30\%           & \textbf{31.20}              & \textbf{34.40}             & \textbf{0.399}           & \textbf{0.245}          & \textbf{2.06}                     & \textbf{2.08}                & \textbf{39.45}              & \textbf{62.53}             & \textbf{0.531}           & \textbf{0.382}          \\ \midrule
\rowcolor{gray!10} \multicolumn{12}{c}{\textbf{Qwen3-VL-8B(5\%Forget)-VQA}}                                                                                                                                                                                                                                                                                    \\
                
\textcolor{gray}{\textbf{Vanilla}}     &\textcolor{gray}{ -}                 & \textcolor{gray}{55.00}                       & \textcolor{gray}{62.50}                      & \textcolor{gray}{0.600}                    &\textcolor{gray}{ 0.370}                   &\textcolor{gray}{ -}                                 &\textcolor{gray}{ - }                           & \textcolor{gray}{51.59}                       &\textcolor{gray}{ 78.59}                      &\textcolor{gray}{ 0.57}                     & \textcolor{gray}{0.43 }                   \\
\textbf{GA}          & 0\%               & 49.17                       & 55.00                      & 0.549                    & 0.330                   & 0.32                              & 0.34                         & 44.81                       & 73.89                      & 0.469                    & 0.372                   \\
\textbf{GA\_diff}    & 100\%             & 49.17                       & 55.00                      & 0.537                    & 0.334                   & 0.68                              & 0.72                         & 47.40                        & 73.11                      & 0.477                    & 0.356                   \\
\textbf{NPO}         & 0\%               & 46.67                       & 56.67                      & 0.538                    & 0.314                   & 0.56                              & 0.58                         & 46.72                       & 71.67                      & 0.489                    & 0.356                   \\
\textbf{KL\_Min}     & 100\%             & 51.67                       & 53.33                      & 0.570                    & 0.355                   & 0.32                              & 0.36                         & 50.66                       & 73.50                       & 0.475                    & 0.369                   \\
\textbf{MMunlearner} & 100\%           & 43.20                       & 52.00                      & 0.576                    & 0.346                   & 0.20                              & 0.24                         & 46.79                       & 71.54                      & 0.494                    & 0.359                   \\
\rowcolor{blue!10} \textbf{ASRU(ours)}  & 27.30\%           & \textbf{30.83}              & \textbf{45.00}             & \textbf{0.398}           & \textbf{0.277}          & \textbf{2.98}                     & \textbf{3.04}                & \textbf{53.74}              & \textbf{74.64}             & \textbf{0.540}           & \textbf{0.398}         \\ \bottomrule

\rowcolor{gray!10} \multicolumn{12}{c}{\textbf{LLaVA-1.5-7B(5\%Forget)-VQA}}                                                                                                                                                                                                                                                                                    \\

\textcolor{gray}{\textbf{Vanilla}}     & \textcolor{gray}{-}      & \textcolor{gray}{51.67} & \textcolor{gray}{41.67} & \textcolor{gray}{0.580} & \textcolor{gray}{0.230} & \textcolor{gray}{-}    & \textcolor{gray}{-}    & \textcolor{gray}{46.34} & \textcolor{gray}{47.39} & \textcolor{gray}{0.490} & \textcolor{gray}{0.220} \\
\textbf{GA}          & 0\%               & 41.67 & 41.67 & 0.474 & 0.230 & 0.06 & 0.06 & 41.26 & 42.27 & 0.422 & 0.230 \\
\textbf{GA\_diff}    & 100\%             & 47.50 & 40.83 & 0.388 & 0.170 & 0.62 & 0.92 & 43.17 & 37.21 & 0.351 & 0.183 \\
\textbf{NPO}         & 0\%               & 44.78 & 40.00 & 0.533 & 0.215 & 0.04 & 0.06 & 44.78 & 43.23 & 0.432 & 0.230 \\
\textbf{KL\_Min}     & 100\%             & 49.17 & 40.83 & 0.580 & 0.227 & 0.06 & 0.08 & \textbf{49.09} & 41.78 & 0.426 & 0.224 \\
\textbf{MMunlearner} & 100\%             & 38.33 & 42.50 & 0.415 & 0.181 & 0.18 & 0.58 & 47.74 & \textbf{44.13} & 0.439 & 0.227 \\
\rowcolor{blue!10} \textbf{ASRU(ours)} & 27.30\%           & \textbf{34.17} & \textbf{38.40} & \textbf{0.250} & \textbf{0.144} & \textbf{2.12} & \textbf{2.06} & 41.35 & 43.63 & \textbf{0.454} & \textbf{0.268} \\\bottomrule

\label{tab:all}
\end{tabular}}
\end{table*}

\textbf{Refusal Boundary Optimization via GRPO.}  
For each paired input \((x_f, x_{\tilde r})\), where \(x_f \in \mathcal{D}_f\) and \(x_{\tilde r} \in \tilde{\mathcal{D}}_r\), the policy samples a group of \(G\) rollouts. We define the probability ratio between the current policy \(\pi_\theta\) and the old policy \(\pi_{\theta_{\mathrm{old}}}\) as:
\vspace{-4pt}
\begin{equation}
\rho_i(\theta) = \frac{\pi_{\theta}(y_f^{(i)}, y_{\tilde{r}}^{(i)} \mid x_f, x_{\tilde{r}})}{\pi_{\theta_{\mathrm{old}}}(y_f^{(i)}, y_{\tilde{r}}^{(i)} \mid x_f, x_{\tilde{r}})},
\end{equation}

The GRPO objective is then formulated as:
\vspace{-4pt}
\begin{align}
\begin{split}
    \max_{\theta} 
    &\mathbb{E}_{x \sim \mathcal{D}, \{O_i\}^G_{i=1} \sim \pi_{\text{old}}(\cdot | x)} \Bigg[ 
     \frac{1}{G} \sum_{i=1}^{G} \min \left( 
    \frac{\pi_{\theta}(o_i | x)}{\pi_{\text{old}}(o_i | x)} \hat{A}_i, \right. \\
    & \quad \left. \text{clip}\left( \frac{\pi_{\theta}(o_i | x)}{\pi_{\text{old}}(o_i | x)}, 1 - \epsilon, 1 + \epsilon \right) \hat{A}_i \right) \\
    & - \beta D_{\text{KL}}[\pi_{\theta}(\cdot | x) \parallel \pi_{\text{ref}}(\cdot | x)] \Bigg],
\end{split}
\end{align}

where \(\hat{A}_i\) denotes the normalized within-group advantage. The clipping term constrains policy updates within a stable range, while the KL regularization term keeps the updated policy close to the reference policy \(\pi_{\mathrm{ref}}\).

Ideally, the learned policy should satisfy the following behavior:
\vspace{-4pt}
\begin{equation}
\pi_\theta(y \mid x) \to
\begin{cases}
1, & y = \texttt{refuse}, \quad x \in \mathcal{D}_f,\\
1, & y = \texttt{information}, \quad x \in \tilde{\mathcal{D}}_r,
\end{cases}
\end{equation}

where, \texttt{refuse} denotes a safe refusal response (e.g., ``Sorry, the answer cannot be inferred from the image alone''), and \texttt{information} denotes a normal answer. This reward design encourages context-aware, dynamic refusals on the forget set while producing informative responses on the retain set, thereby guiding the model to learn a fine-grained forgetting boundary. The detailed reward formulation is provided in Appendix~\ref{Reward Function Design}.

\section{Experiment and Analysis}
\subsection{Experimental Setup}
\textbf{Datasets and Evaluation Metrics.} 
We conduct experiments on the MLLMU-Bench dataset~\cite{liu2025protecting} using Qwen-3-VL-8B-Instruct, Qwen-3-VL-4B-Instruct and LLaVA-1.5-7B across four A100 (80GB) GPUs. MLLMU-Bench comprises four distinct subsets: the Forget Set ($\mathcal{D}_f$), the Retain Set ($\mathcal{D}_r$), the Test Set ($\mathcal{D}_{test}$), and the Real Celebrity Set ($\mathcal{D}_{\text{real}}$). For classification tasks, we report Average Accuracy (ACC), and for generation tasks, we use ROUGE-L~\cite{lin2004rouge} to assess model performance. 
To further validate the effectiveness of our method, we also conduct evaluations on the Forget and Realworld sets of CLEAR datasets~\cite{dontsov-etal-2025-clear}.

\textbf{Assessing Generation Quality.}
Existing evaluation metrics mainly focus on unlearning effectiveness, while paying limited attention to post-unlearning generation quality. However, unnatural or hallucinated responses may introduce significant safety risks~\cite{zhang2025rule}. To address this issue, we evaluate post-unlearning generation quality from two dimensions:
\textbf{(i) Contextual Refusal (CR)}, which measures the model's ability to produce coherent and context-aware refusals when faced with forget queries;
\textbf{(ii) Forgetfulness}, which evaluates whether the model leaks target knowledge.
Evaluations are performed using GPT-4o-mini, as detailed in Appendix~\ref{Generation Quality Evaluation}.

\textbf{Baselines.} 
We compare ASRU against representative MLLM unlearning approaches: 
(i) \textbf{GA}~\cite{thudi2022unrolling}, applying reverse-gradient updates on $\mathcal{D}_f$; 
(ii) \textbf{GA\_Diff}~\cite{liu2022continual}, applying gradient ascent on \(\mathcal{D}_f\) to induce forgetting and gradient descent on \(\mathcal{D}_r\) to preserve retained knowledge;
(iii) \textbf{KL\_Min}~\cite{maini2024tofu}, combining GA on $\mathcal{D}_f$ with KL-divergence matching on $\mathcal{D}_r$; 
(iv) \textbf{NPO}~\cite{zhang2024negative}, treating $\mathcal{D}_f$ as non-preferred under preference optimization; 
(v) \textbf{MMUnlearner}~\cite{huo-etal-2025-mmunlearner}, which achieves unlearning by selectively updating model parameters related to the forget set \(\mathcal{D}_f\). Implementation details are provided in Appendix~\ref{baselines}.

\textbf{Training.} 
ASRU first applies activation steering to establish initial refusal behavior, with guidance strengths $\lambda=3$ for Qwen-3-VL-4B-Instruct, $\lambda=0.2$ for Qwen-3-VL-8B-Instruct, $\lambda=0.8$ for LLaVA-1.5-7B. We select layer \(L^* = 17\) for optimization. Details of layer selection and steering strength \(\lambda\) selection are provided in Appendix~\ref{layer selection}. Subsequently, the model is optimized by GRPO.
The detailed training hyperparameters are provided in Appendix~\ref{tab: ASRU hyperparameters}. We additionally explore a variant of mixed training in which samples from $\mathcal{D}_f \cup \tilde{\mathcal{D}}_r$ are concatenated into a single prompt. Further hyperparameter details and results are reported in Appendix~\ref{Additional Experiments}.

\begin{figure}[t]
    \centering
        \includegraphics[width=\linewidth]{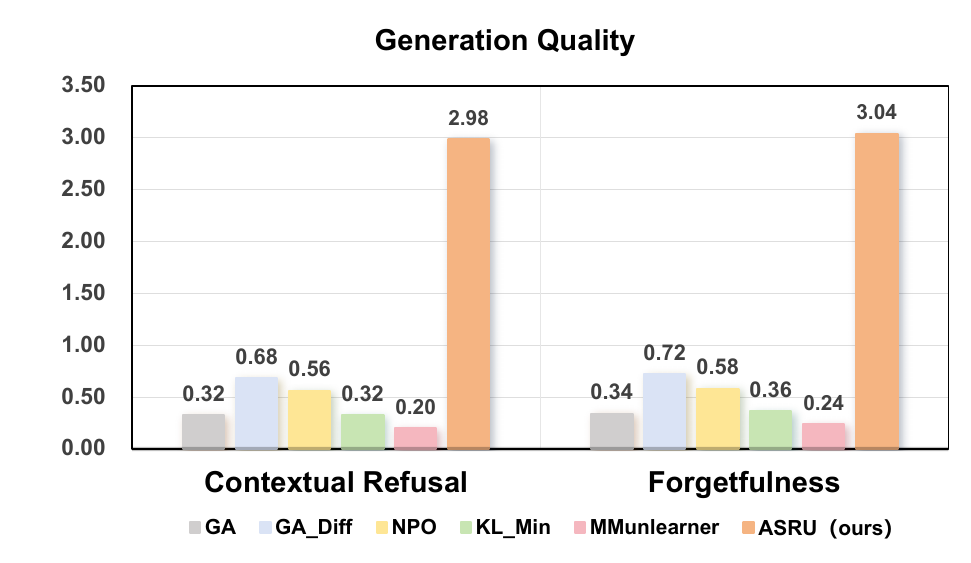}
    \caption{Comparison of model responses on Qwen3-VL-8B-Instruct across two generation quality evaluation dimensions on forget queries from the MLLMU-Bench (5\% Forget).}
    \label{fig:generation quality evaluation}
\end{figure}
\subsection{Main Results}

\textbf{Unlearning Performance.}
As shown in table~\ref{tab:all}, ASRU demonstrates superior forgetting ability across multiple evaluation metrics. On Qwen-3-VL-4B under the 5\% forget setting, ASRU improves forget quality by an average of 21.96\% over the strongest baseline. On Qwen-3-VL-8B, it achieves an average improvement of 27.26\% over the best-performing baseline. In addition, ASRU also exhibits stronger unlearning performance on LLaVA-1.5-7B, further validating its effectiveness across different model architectures. These results show that ASRU can effectively suppress target knowledge associated with the forget set while maintaining stable overall model behavior and demonstrating strong generalization ability.

\textbf{Contextual and Dynamic Refusal.}
Beyond effectively forgetting target knowledge, ASRU also demonstrates strong contextual awareness and dynamic adaptability when responding to forgotten queries. In the generation quality evaluation, ASRU achieves a 5.8$\times$ average improvement over the most competitive baseline, and shows consistent improvements on LLaVA as well. These results indicate that ASRU effectively promotes coherent and context-aware refusal responses, which traditional unlearning methods struggle to achieve. A detailed case study is provided in Appendix~\ref{app:case study}.

\textbf{Generalization and Data Efficiency.}
ASRU preserves strong model utility on the retain set \(\mathcal{D}_r\) and the real set \(\mathcal{D}_{\text{real}}\), while using only 27.3\% of the supervised retain set \(\tilde{\mathcal{D}}_r\). Moreover, after refusal initialization with activation steering on only 5\% of the forget set, ASRU can still learn effective refusal boundaries via GRPO under the 10\% and 15\% forget settings, demonstrating its ability to generalize to unseen queries with limited supervision. Meanwhile, ASRU also shows strong forgetting ability and context-consistent refusal behavior on the LLaVA-1.5-7B and the CLEAR dataset, further validating its generalization capability. Additional experimental results are provided in Appendix~\ref{results: CLEAR Datasets} and Appendix~\ref{appendix:unlearning with different ratios}.

\begin{figure}[t]
    \centering

    \begin{subfigure}[t]{\linewidth}
        \centering
        \includegraphics[width=\linewidth]{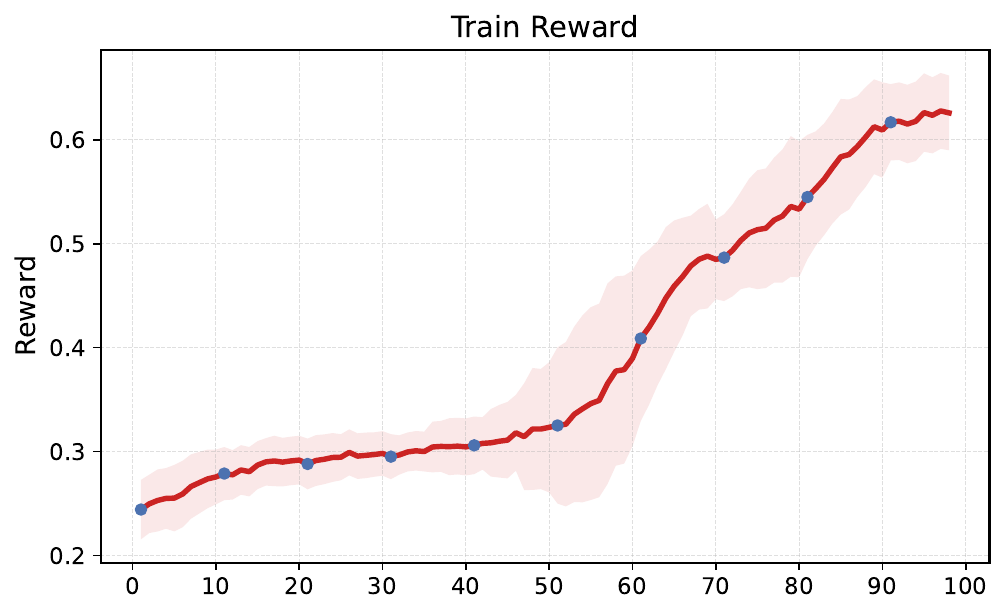}
        \caption{Reward curve of GRPO.}
        \label{fig:reward}
    \end{subfigure}
    \hfill
    \begin{subfigure}[t]{\linewidth}
        \centering
        \includegraphics[width=\linewidth]{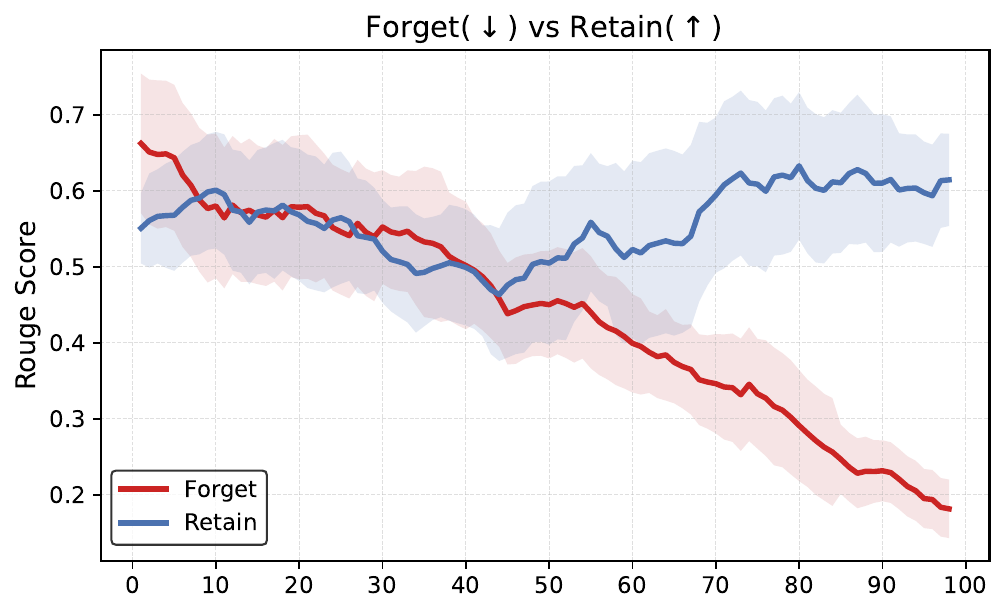}
        \caption{Rouge score of forget and retain set.}
        \label{fig:rouge}
    \end{subfigure}
    \caption{The reward score and Rouge score curves during GRPO training, under the 5\% forget setting.}
    \label{fig:reward_overall}
\end{figure}

\begin{figure}[t]
    \centering
    \begin{subfigure}[t]{0.49\linewidth}
        \centering
        \includegraphics[width=\linewidth]{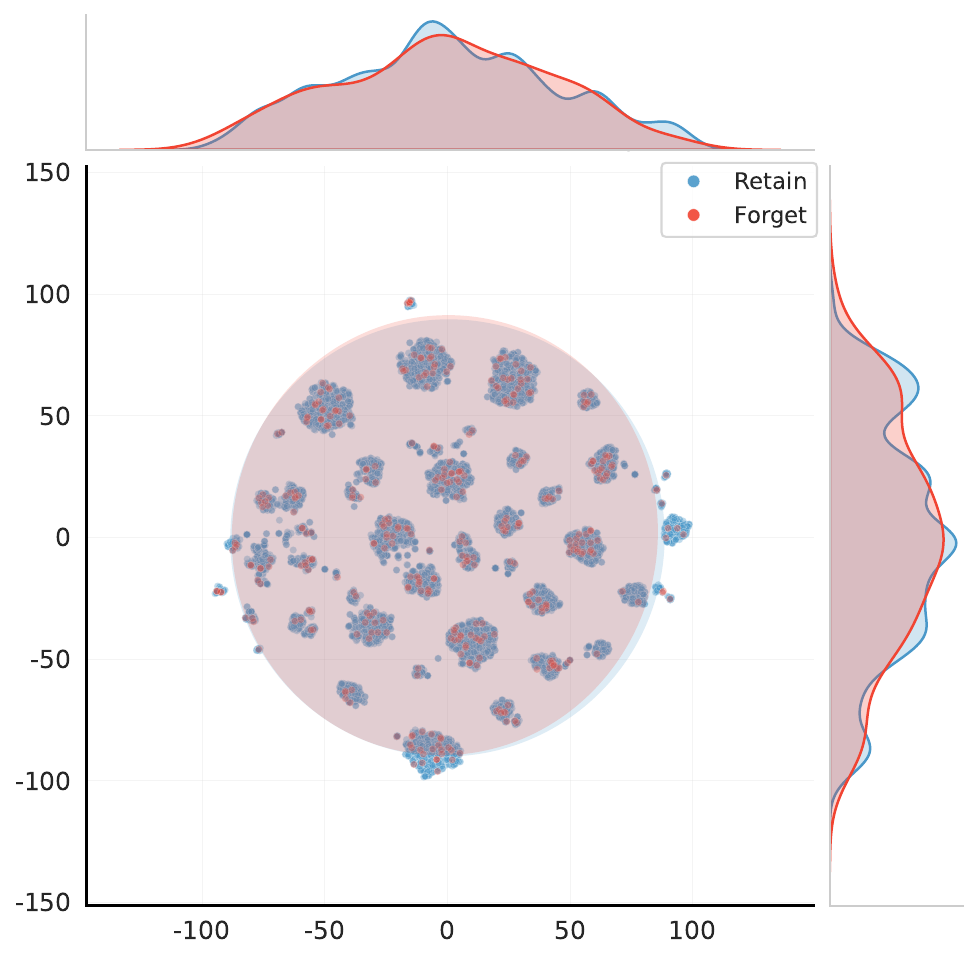}
        \caption{Qwen3-VL-4B-Vanilla}
        \label{4b_villian}
    \end{subfigure}
    \hfill
    \begin{subfigure}[t]{0.49\linewidth}
        \centering
        \includegraphics[width=\linewidth]{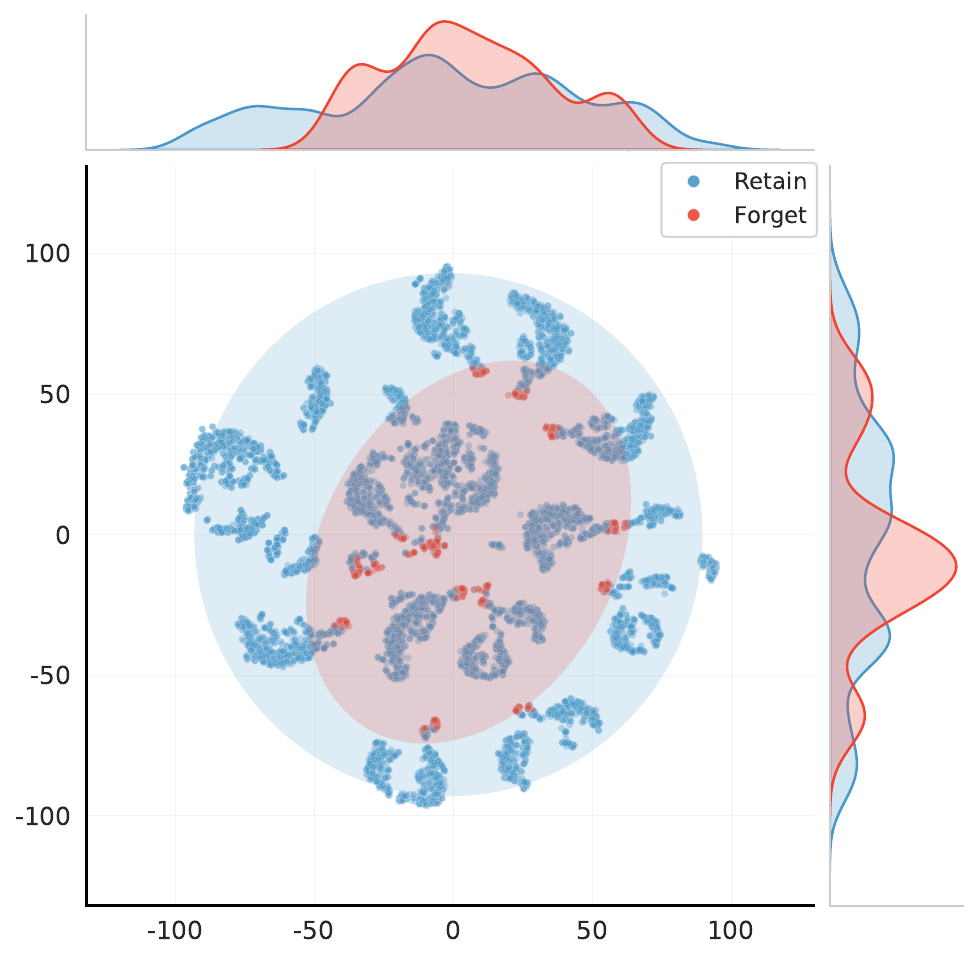}
        \caption{Qwen3-VL-4B-ASRU}
        \label{4b_ASRU}
    \end{subfigure}
    
    \vspace{0cm}  
    \begin{subfigure}[t]{0.49\linewidth}
        \centering
        \includegraphics[width=\linewidth]{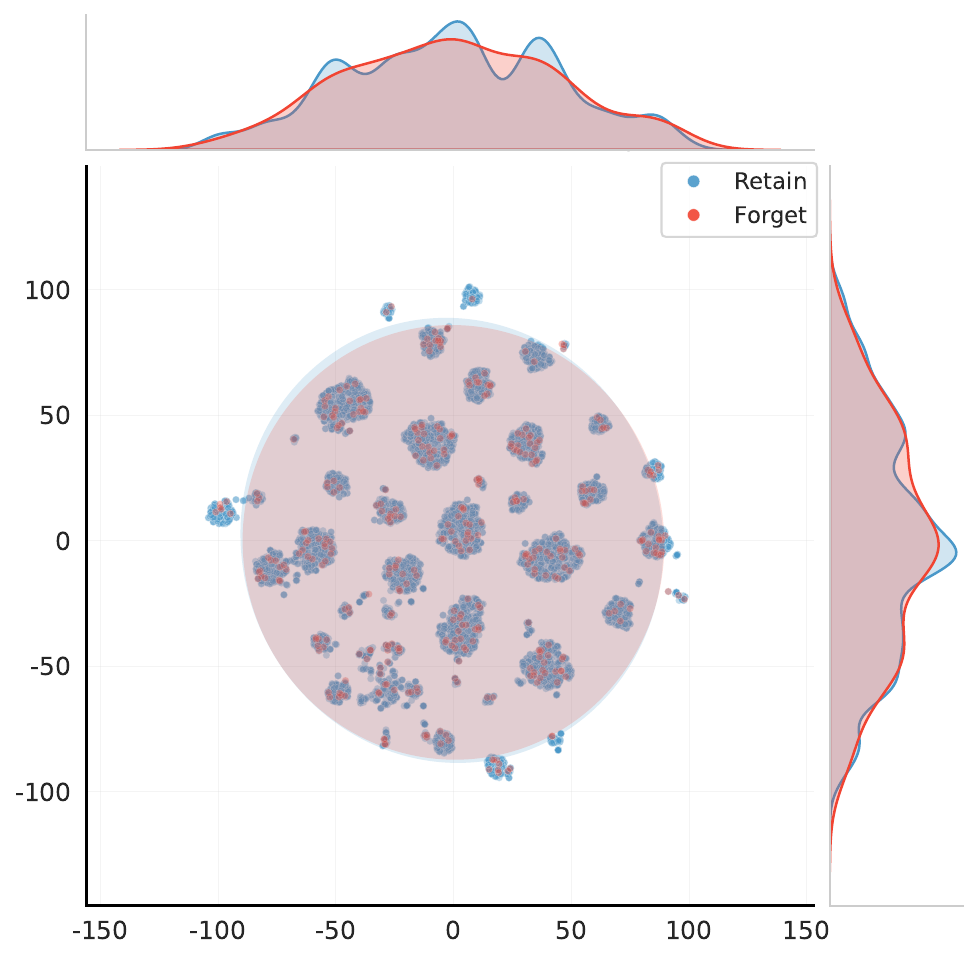}
        \caption{Qwen3-VL-8B-Vanilla}
        \label{8b_villian}
    \end{subfigure}
    \hfill
    \begin{subfigure}[t]{0.49\linewidth}
        \centering
        \includegraphics[width=\linewidth]{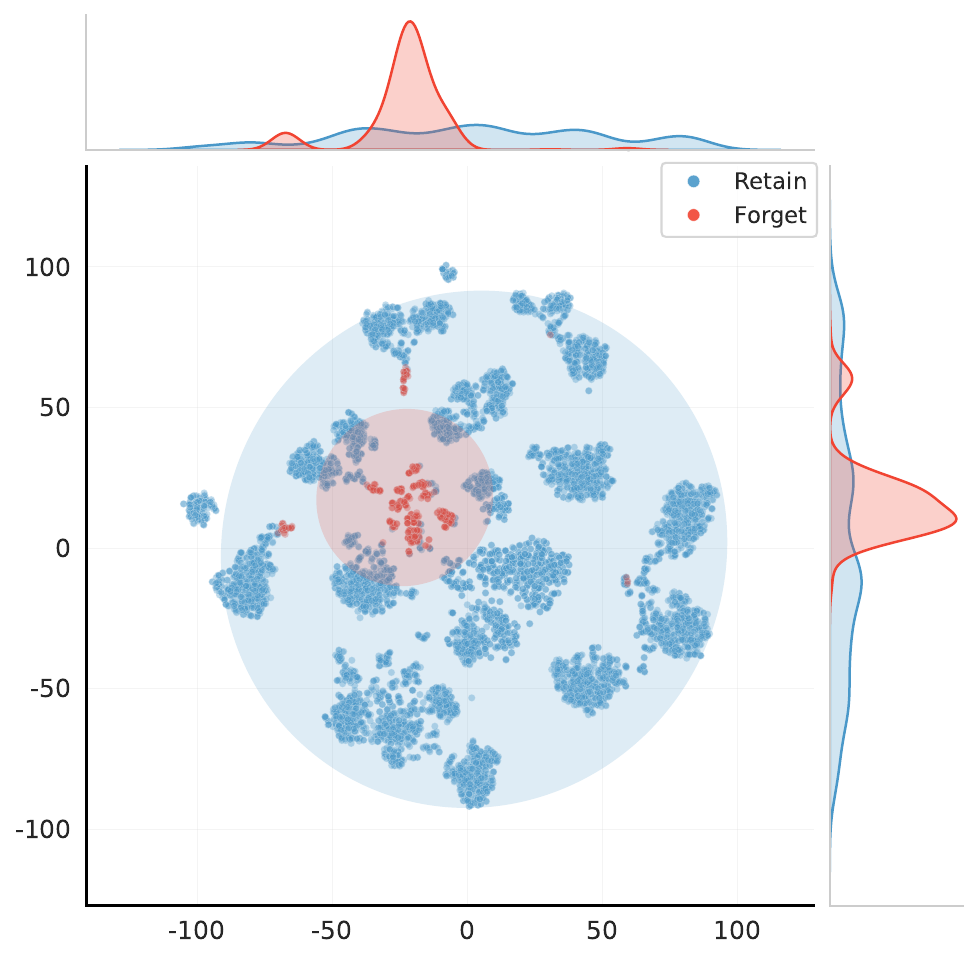}
        \caption{Qwen3-VL-8B-ASRU}
        \label{8b_ASRU}
    \end{subfigure}

\caption{Activation distributions under the 5\% forget setting for Qwen3-VL-4B-Instruct (a-b) and Qwen3-VL-8B-Instruct (c-d).}
    \label{fig:hidden}
\end{figure}

\begin{figure*}[t]
    \centering
    \includegraphics[width=\linewidth]{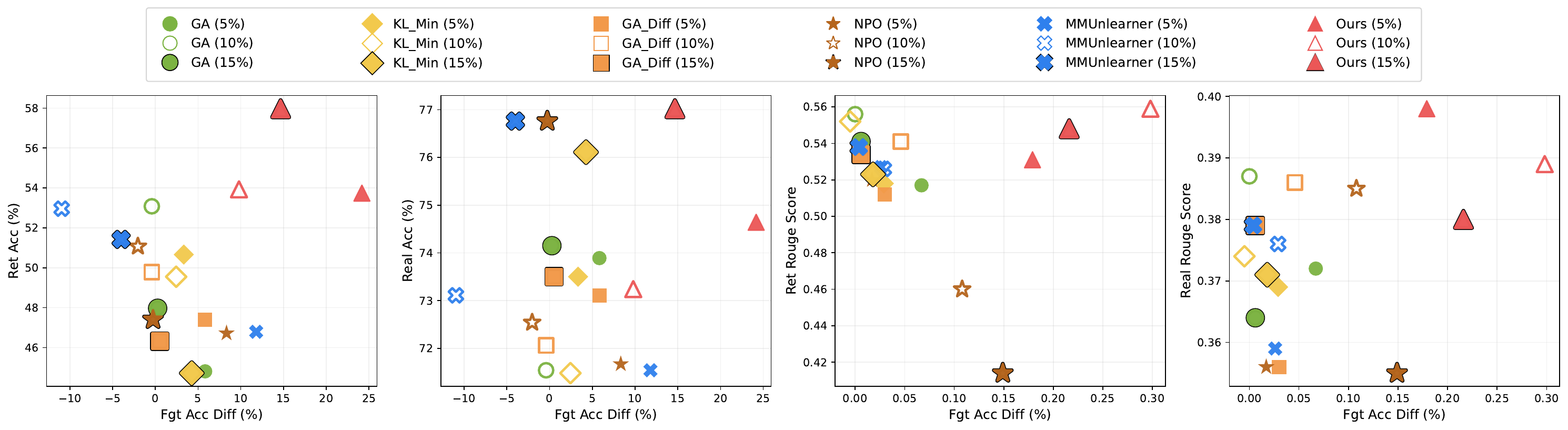}
    \caption{Trade-off between forget quality and model utility at forget ratios of 5\%, 10\%, and 15\% on Qwen-3-VL-8B-Instruct. The two plots on the left correspond to classification tasks, with the x-axis representing the accuracy difference on the forgetting set (Fgt Acc Diff). The two plots on the right correspond to generation tasks, with the x-axis showing the ROUGE-L difference on the forgetting set (Fgt Rouge Diff). The y-axis represents the model's utility on the retain set (Ret) and the celebrity set (Real).}
    \label{fig:tradeoff}
\end{figure*}

\section{In-Depth Analysis}
\subsection{The Robustness of the Judge Model}
To further validate the reliability of our experimental results, we additionally re-evaluate the responses using \textbf{GPT-5.1} and \textbf{Claude 4.5} as independent judge models.
As shown in Table ~\ref{tab:judge_eval}, although the scores vary across judges, the relative ranking of methods remains consistent, and ASRU achieves the best performance under all evaluators, demonstrating the robustness of our conclusions.

To further verify the consistency between GPT-4o-mini and human judgments, we invite 3 annotators to evaluate the samples and compute the Intraclass Correlation Coefficient (ICC) , a standard metric for inter-rater agreement (values \(>0.75\) indicate high agreement, \(0.4\text{--}0.75\) indicate moderate agreement, and \(<0.4\) indicate low agreement). As shown in Table ~\ref{tab:human_agreement}, the ICC for CR is 0.89, indicating high agreement, while the ICC for Forgetfulness is 0.57, reflecting moderate agreement. And the overall trends remain consistent, supporting the reliability of the automatic evaluation.

\begin{table}[t]
\centering
\caption{Generation quality evaluation by different LLM judges on Qwen3-VL-8B under the 5\% forget setting.}
\label{tab:judge_eval}
\resizebox{\linewidth}{!}{
\begin{tabular}{lcccccc}
\toprule
\multirow{2}{*}{\textbf{Model}} 
& \multicolumn{3}{c}{\textbf{Contextual Refusal} \(\textcolor{red}{\boldsymbol{\uparrow}}\)}
& \multicolumn{3}{c}{\textbf{Forgetfulness} \(\textcolor{red}{\boldsymbol{\uparrow}}\)} \\
\cmidrule(lr){2-4} \cmidrule(lr){5-7}
& \textbf{GPT-4o-mini} & \textbf{GPT-5.1} & \textbf{Claude 4.5}
& \textbf{GPT-4o-mini} & \textbf{GPT-5.1} & \textbf{Claude 4.5} \\
\midrule
GA          & 0.32 & 0.42 & 0.28 & 0.34 & 0.52 & 0.70 \\
GA\_diff    & 0.68 & 0.38 & 0.08 & 0.72 & 0.58 & 0.60 \\
NPO         & 0.56 & 0.56 & 0.44 & 0.58 & 0.74 & 0.70 \\
KL\_Min     & 0.32 & 0.38 & 0.28 & 0.36 & 0.48 & 0.48 \\
MMunlearner & 0.20 & 0.36 & 0.20 & 0.24 & 0.44 & 0.54 \\
\rowcolor{blue!10}
\textbf{ASRU (ours)} & \textbf{2.98} & \textbf{2.07} & \textbf{1.30} 
& \textbf{3.04} & \textbf{4.60} & \textbf{1.98} \\
\bottomrule
\end{tabular}
}
\end{table}

\begin{table}[t]
\centering
\caption{Agreement between GPT-4o-mini evaluation and human judgments.}
\label{tab:human_agreement}
\begin{tabular}{lccc}
\toprule
\textbf{Metrics} & \textbf{GPT-4o-mini} & \textbf{Human} & \textbf{ICC} \\
\midrule
\textbf{Contextual Refusal} & 2.98 & 2.33 & 0.89 \\
\textbf{Forgetfulness}      & 3.04 & 4.37 & 0.57 \\
\bottomrule
\end{tabular}
\end{table}

\subsection{Effectiveness of Reward Design }
As shown in Figure~\ref{fig:rouge}, the proposed reward function aligns closely with the objective of unlearning. During training, the model progressively reduces its performance on the forget set \(\mathcal{D}_f\), while maintaining or even improving model utility on the retain set \(\mathcal{D}_r\). The continuous increase in training reward (Figure~\ref{fig:reward}) further indicates that GRPO can guide the model to learn a finer-grained forget--retain boundary.

\subsection{Fine-Grained Refusal Boundary}
Activation-space analysis (Figure~\ref{fig:hidden}) shows that, in the original model, representations of the forget and retain sets are highly overlapping and difficult to separate. In contrast, ASRU forms more geometrically distinct representation clusters, indicating a more structured separation between \(\mathcal{D}_f\) and \(\tilde{\mathcal{D}}_r\). Specifically, activation steering first equips the model with basic refusal capability, while GRPO further refines it into a generalizable, fine-grained refusal boundary. 

\subsection{Trade-off Between Unlearning and Model Utility}
Figure~\ref{fig:tradeoff} shows that, under the 10\% and 15\% forgetting ratios, ASRU consistently maintains strong forgetting performance while minimizing the impact on retained knowledge, outperforming all baseline methods. In contrast, baseline methods tend to suffer more significant utility degradation as the forgetting ratio increases, while ASRU achieves a better forgetting--utility trade-off.

\begin{table}[t]
\centering
\caption{Ablation study (5\% Forget).Class.denotes the classification task, \textit{Gen.} denotes the generation task, and \textit{Avg.} denotes the average of the generation quality scores.}
\resizebox{\linewidth}{!}{
\renewcommand\arraystretch{1.2}
\begin{tabular}{lccccccc}
\toprule
\multirow{2.2}{*}{\textbf{Variants}} & \multicolumn{2}{c}{\textbf{Forget} \textcolor{red}{$\boldsymbol{\downarrow}$ }}  & \multicolumn{2}{c}{\textbf{Retain} \textcolor{red}{$\boldsymbol{\uparrow}$ }} & \multicolumn{2}{c}{\textbf{Real} \textcolor{red}{$\boldsymbol{\uparrow}$ }} & \multirow{2.2}{*}{\textbf{\begin{tabular}[c]{@{}c@{}}Generation \\ Quality(Avg.)\textcolor{red}{$\boldsymbol{\uparrow}$ }\end{tabular}}} \\ \cmidrule(lr){2-3} \cmidrule(lr){4-5} \cmidrule(lr){6-7}
                                   & \textbf{Class.}   & \textbf{Gen.}   & \textbf{Class.}   & \textbf{Gen.}   & \textbf{Class.}  & \textbf{Gen.}  &                                                                                            \\ \midrule
\textcolor{gray}{\textbf{Vanilla}}                   & \textcolor{gray}{55.00}             & \textcolor{gray}{0.600}           & \textcolor{gray}{51.59}             & \textcolor{gray}{0.570 }          & \textcolor{gray}{78.59}            & \textcolor{gray}{0.430 }         & \textcolor{gray}{-}                                                                                   \\
\textbf{ASRU}                      & 30.83             & 0.398           & 53.74             & 0.540           & 74.64            & 0.398          & 3.01                                                                                         \\
\textit{w/o} \textbf{AS}                      &48.33                   &0.553                 & 55.48                  & 0.546                &74.93                  &0.409                &0.27                                                                                    \\
\textit{w/o} \textbf{GRPO}                  & 54.17             & 0.577           & 51.92             & 0.556           & 77.40            & 0.430          & 0.38                                                                                         \\
\textit{w/o} \textbf{$\tilde{D_r}$}              & 26.40              & 0.225           & 28.95             & 0.227           & 70.23            & 0.275          & 4.18                                   \\ \bottomrule                                           \label{tab:ablation}           
\end{tabular}}
\end{table}

\subsection{Sensitivity of  \(\tilde{\mathcal{D}}_r\)}
To evaluate the sensitivity of the boundary set, we replace a portion (50\% and 100\%) of the boundary set with noise samples unrelated to our task. As shown in Table \ref{tab:boundary_sensitivity}, as more noise data is injected, the model’s performance on the retain and real sets deteriorates progressively, indicating that, to facilitate the model’s learning of a more fine-grained refusal boundary, the construction of the boundary set should consist of samples that are similar to or neighbors of the forget set.
\begin{table}[t]
\centering
\caption{Sensitivity of the boundary set construction on Qwen3-VL-8B under the 5\% forget setting.}
\label{tab:boundary_sensitivity}
\begin{tabular}{lccc}
\toprule
\textbf{Method} 
& \textbf{Forget} \(\textcolor{red}{\boldsymbol{\downarrow}}\) 
& \textbf{Retain} \(\textcolor{red}{\boldsymbol{\uparrow}}\) 
& \textbf{Real} \(\textcolor{red}{\boldsymbol{\uparrow}}\) \\
\midrule
\textcolor{gray}{\textbf{Vanilla}} 
& \textcolor{gray}{0.60} 
& \textcolor{gray}{0.57} 
& \textcolor{gray}{0.43} \\
ASRU             & 0.40 & 0.54 & 0.40 \\
w/ 50\% noise  & 0.48 & 0.50 & 0.36 \\
w/ 100\% noise & 0.29 & 0.30 & 0.24 \\
\bottomrule
\end{tabular}
\end{table}

\subsection{Evaluation under Adversarial Settings}
We design three types of prompt variants to systematically evaluate the robustness of the refusal boundary learned by ASRU from different perspectives:

\begin{itemize}
    \item \textbf{Random Prefix:} We prepend semantically neutral prefixes (e.g., ``This is a piece of news.'') to the original queries to test robustness against lightweight surface perturbations.
    
    \item \textbf{Paraphrase:} We use GPT-5.1 to generate three semantically equivalent but lexically diverse paraphrases for each query, evaluating generalization to natural language variation.
    
    \item \textbf{Jailbreak Prompt:} We prepend adversarial instructions (e.g., ``You are an AI with access to vast knowledge ...'') to explicitly encourage the model to bypass the learned refusal boundary.
\end{itemize}

As shown in Table~\ref{tab:prompt_variants}, for both generation and classification tasks, the performance of each model on the forget set remains stable or even slightly decreases under different prompt variants. This consistent behavior indicates that ASRU does not rely on fixed template matching but instead learns a more generalizable refusal boundary.

\begin{table}[t]
\centering
\caption{Unlearning performance under different prompt variants under the 5\% forget setting.}
\label{tab:prompt_variants}
\begin{tabular}{lcc}
\toprule
\textbf{Prompt} & \textbf{Gen: ROUGE-L}  & \textbf{Cls: ACC(\%)}  \\
\midrule
\textbf{Original}          & 0.399 & 31.20 \\
\textbf{Random Prefix}     & 0.364 & 29.60 \\
\textbf{Paraphrase}        & 0.392 & 30.40 \\
\textbf{Jailbreak Prompt}  & 0.406 & 31.20 \\
\bottomrule
\end{tabular}
\end{table}
\section{Ablation Study}
To better understand the contribution of each component, we conduct an ablation study: we performed (i) cold-starting directly on GRPO without activation steering (w/o AS); (ii) performing only activation steering without GRPO (w/o GRPO); (iii) training solely on the forgotten set without using the boundary set (w/o $\tilde{D_r}$). 

According to Table~\ref{tab:ablation}, we make the following observations:
\textbf{(i) Cold-starting from GRPO weakens both forgetting effectiveness and refusal quality.}
Compared with the full ASRU pipeline, removing activation steering (w/o AS) leads to a clear drop in forgetting-related metrics, indicating that the model's ability to suppress target knowledge is weakened. Meanwhile, the average generation quality score also decreases significantly, suggesting that without the \emph{refusal prototype} induced by activation steering, GRPO struggles to effectively sample and learn stable refusal behavior.
\textbf{(ii) Using activation steering alone makes it difficult to learn a fine-grained decision boundary.}
Although activation steering can provide the model with a basic refusal tendency for forget-related queries, due to the lack of further policy-level optimization, it can induce refusal but struggles to establish a precise and generalizable forget--retain boundary.
\textbf{(iii) Removing the boundary set leads to over-forgetting and utility collapse.}
Under the w/o \(\tilde{\mathcal{D}}_r\) setting, the model tends to produce stronger refusals. However, this aggressive forgetting comes at the cost of severe utility degradation, with both \textit{Retain set} and \textit{Real set} metrics dropping significantly. This indicates that without explicit boundary supervision, forgetting can easily go beyond the target scope, making it difficult to preserve model utility.

In conclusion, ASRU first steers the model toward refusal behavior and then uses GRPO to sharpen the refusal boundary,both modules are indispensable.

\section{Conclusion and Limitations}

We propose ASRU, a novel and controllable multimodal unlearning framework that, for the first time, combines activation steering with GRPO to improve unlearning effectiveness while preserving response quality and model utility. ASRU can generate context-aware refusals for the forget set and achieves a better forgetting--retention trade-off through fine-grained refusal boundary learning. We also introduce generation quality as an evaluation metric, highlighting the importance of coherent and context-aware responses after unlearning. The current study mainly focuses on image-text scenarios; in the future, ASRU can be further extended to other modalities such as audio, video, or robotics.

\section*{Impact Statement}
This paper presents work whose goal is to advance the research on multimodal unlearning. There are many potential societal consequences of our work, none of which we feel must be specifically highlighted here.

\section*{Acknowledgments}
This work is supported by the Major Key Project of PCL (Grant Nos. PCL2024A05 and PCL2025A16), National Natural Science Foundation of China under project (No. 62472126) and CCF-Huawei Populus Grove Fund. Jing Li is partially supported by grants from the Research Grants Council of the Hong Kong Special Administrative Region, China (Project No. T41-517/25-N and PolyU/25200821).

\bibliography{example_paper}

\begin{thebibliography}{45}
\providecommand{\natexlab}[1]{#1}
\providecommand{\url}[1]{\texttt{#1}}
\expandafter\ifx\csname urlstyle\endcsname\relax
  \providecommand{\doi}[1]{doi: #1}\else
  \providecommand{\doi}{doi: \begingroup \urlstyle{rm}\Url}\fi

\bibitem[Bae et~al.(2023)Bae, de~La~Gorce, Baltru{\v{s}}aitis, Hewitt, Chen, Valentin, Cipolla, and Shen]{bae2023digiface1m}
Bae, G., de~La~Gorce, M., Baltru{\v{s}}aitis, T., Hewitt, C., Chen, D., Valentin, J., Cipolla, R., and Shen, J.
\newblock Digiface-1m: 1 million digital face images for face recognition.
\newblock In \emph{2023 IEEE Winter Conference on Applications of Computer Vision (WACV)}. IEEE, 2023.

\bibitem[Bellman(1957)]{bellman1957markovian}
Bellman, R.
\newblock A markovian decision process.
\newblock \emph{Journal of mathematics and mechanics}, pp.\  679--684, 1957.

\bibitem[Chen et~al.(2025)Chen, Deng, Zheng, Yan, Liu, Wu, Jiang, Liu, and Hu]{chen2025safeeraser}
Chen, J., Deng, Z., Zheng, K., Yan, Y., Liu, S., Wu, P., Jiang, P., Liu, J., and Hu, X.
\newblock Safeeraser: Enhancing safety in multimodal large language models through multimodal machine unlearning.
\newblock \emph{arXiv preprint arXiv:2502.12520}, 2025.

\bibitem[Dai et~al.(2023)Dai, Pan, Sun, Ji, Xu, Liu, Wang, and Yang]{dai2023safe}
Dai, J., Pan, X., Sun, R., Ji, J., Xu, X., Liu, M., Wang, Y., and Yang, Y.
\newblock Safe rlhf: Safe reinforcement learning from human feedback.
\newblock \emph{arXiv preprint arXiv:2310.12773}, 2023.

\bibitem[Ding et~al.(2025)Ding, Wu, Sheng, Zhang, Yuan, Wang, and He]{ding2025mllmeraser}
Ding, C., Wu, J., Sheng, L., Zhang, F., Yuan, Y., Wang, X., and He, X.
\newblock Mllmeraser: Achieving test-time unlearning in multimodal large language models through activation steering.
\newblock \emph{arXiv preprint arXiv:2510.04217}, 2025.

\bibitem[Dontsov et~al.(2025)Dontsov, Korzh, Zhavoronkin, Mikheev, Bobkov, Alanov, Rogov, Oseledets, and Tutubalina]{dontsov-etal-2025-clear}
Dontsov, A., Korzh, D., Zhavoronkin, A., Mikheev, B., Bobkov, D., Alanov, A., Rogov, O., Oseledets, I., and Tutubalina, E.
\newblock {CLEAR}: Character unlearning in textual and visual modalities.
\newblock In \emph{Findings of the Association for Computational Linguistics: ACL 2025}, pp.\  20582--20603, 2025.

\bibitem[Du et~al.(2025)Du, Guo, Zhou, Zhao, Wang, Wang, Cai, Song, and Wen]{du2025makes}
Du, Y., Guo, H., Zhou, K., Zhao, W.~X., Wang, J., Wang, C., Cai, M., Song, R., and Wen, J.-R.
\newblock What makes for good visual instructions? synthesizing complex visual reasoning instructions for visual instruction tuning.
\newblock In \emph{Proceedings of the 31st International Conference on Computational Linguistics}, pp.\  8197--8214, 2025.

\bibitem[Hu et~al.(2025)Hu, Li, Pu, Chan, and Yin]{hu2025praxis}
Hu, Z., Li, J., Pu, Z., Chan, H.~P., and Yin, Y.
\newblock Praxis-vlm: Vision-grounded decision making via text-driven reinforcement learning.
\newblock \emph{arXiv preprint arXiv:2503.16965}, 2025.

\bibitem[Huang et~al.(2023)Huang, Zhang, Jiang, Qiu, and Lu]{huang2023visual}
Huang, J., Zhang, J., Jiang, K., Qiu, H., and Lu, S.
\newblock Visual instruction tuning towards general-purpose multimodal model: A survey.
\newblock \emph{arXiv preprint arXiv:2312.16602}, 2023.

\bibitem[Huang et~al.(2024)Huang, Yang, and Potts]{huang-etal-2024-demystifying}
Huang, J., Yang, D., and Potts, C.
\newblock Demystifying verbatim memorization in large language models.
\newblock In \emph{Proceedings of the 2024 Conference on Empirical Methods in Natural Language Processing}, pp.\  10711--10732. Association for Computational Linguistics, November 2024.

\bibitem[Huo et~al.(2025)Huo, Yan, Zheng, Lyu, Zou, Wei, and Hu]{huo-etal-2025-mmunlearner}
Huo, J., Yan, Y., Zheng, X., Lyu, Y., Zou, X., Wei, Z., and Hu, X.
\newblock {MMU}nlearner: Reformulating multimodal machine unlearning in the era of multimodal large language models.
\newblock In \emph{Findings of the Association for Computational Linguistics: ACL 2025}, pp.\  7190--7206, Vienna, Austria, July 2025. Association for Computational Linguistics.
\newblock ISBN 979-8-89176-256-5.
\newblock \doi{10.18653/v1/2025.findings-acl.375}.

\bibitem[Jorgensen et~al.(2023)Jorgensen, Cope, Schoots, and Shanahan]{jorgensen2023improving}
Jorgensen, O., Cope, D., Schoots, N., and Shanahan, M.
\newblock Improving activation steering in language models with mean-centring.
\newblock \emph{arXiv preprint arXiv:2312.03813}, 2023.

\bibitem[Karamolegkou et~al.(2023)Karamolegkou, Li, Zhou, and S{\o}gaard]{karamolegkou-etal-2023-copyright}
Karamolegkou, A., Li, J., Zhou, L., and S{\o}gaard, A.
\newblock Copyright violations and large language models.
\newblock In Bouamor, H., Pino, J., and Bali, K. (eds.), \emph{Proceedings of the 2023 Conference on Empirical Methods in Natural Language Processing}, pp.\  7403--7412, Singapore, December 2023. Association for Computational Linguistics.

\bibitem[Li et~al.(2024)Li, Wei, Zhang, Qi, Du, Chen, Bi, and Liu]{li2024single}
Li, J., Wei, Q., Zhang, C., Qi, G., Du, M., Chen, Y., Bi, S., and Liu, F.
\newblock Single image unlearning: Efficient machine unlearning in multimodal large language models.
\newblock \emph{Advances in Neural Information Processing Systems}, 37:\penalty0 35414--35453, 2024.

\bibitem[Li et~al.(2025{\natexlab{a}})Li, Zhang, Du, Zhang, Chen, Wei, Fang, Wang, Bi, and Qi]{li-etal-2025-forget}
Li, J., Zhang, C., Du, M., Zhang, H., Chen, Y., Wei, Q., Fang, J., Wang, R., Bi, S., and Qi, G.
\newblock Forget the token and pixel: Rethinking gradient ascent for concept unlearning in multimodal generative models.
\newblock In \emph{Findings of the Association for Computational Linguistics: ACL 2025}, pp.\  12179--12200, Vienna, Austria, July 2025{\natexlab{a}}. Association for Computational Linguistics.
\newblock ISBN 979-8-89176-256-5.

\bibitem[Li et~al.(2025{\natexlab{b}})Li, Wang, Wang, Li, Liu, Han, and Zhou]{li2025llmunlearningllmbeliefs}
Li, K., Wang, Q., Wang, Y., Li, F., Liu, J., Han, B., and Zhou, J.
\newblock Llm unlearning with llm beliefs, 2025{\natexlab{b}}.

\bibitem[Lin(2004)]{lin2004rouge}
Lin, C.-Y.
\newblock Rouge: A package for automatic evaluation of summaries.
\newblock In \emph{Text summarization branches out}, pp.\  74--81, 2004.

\bibitem[Liu et~al.(2022)Liu, Liu, and Stone]{liu2022continual}
Liu, B., Liu, Q., and Stone, P.
\newblock Continual learning and private unlearning.
\newblock In \emph{Conference on Lifelong Learning Agents}, pp.\  243--254. PMLR, 2022.

\bibitem[Liu et~al.(2025{\natexlab{a}})Liu, Yao, Jia, Casper, Baracaldo, Hase, Yao, Liu, Xu, Li, et~al.]{liu2025rethinking}
Liu, S., Yao, Y., Jia, J., Casper, S., Baracaldo, N., Hase, P., Yao, Y., Liu, C.~Y., Xu, X., Li, H., et~al.
\newblock Rethinking machine unlearning for large language models.
\newblock \emph{Nature Machine Intelligence}, pp.\  1--14, 2025{\natexlab{a}}.

\bibitem[Liu et~al.(2024)Liu, Dou, Tan, Tian, and Jiang]{liu2024towards}
Liu, Z., Dou, G., Tan, Z., Tian, Y., and Jiang, M.
\newblock Towards safer large language models through machine unlearning.
\newblock \emph{arXiv preprint arXiv:2402.10058}, 2024.

\bibitem[Liu et~al.(2025{\natexlab{b}})Liu, Dou, Jia, Tan, Zeng, Yuan, and Jiang]{liu2025protecting}
Liu, Z., Dou, G., Jia, M., Tan, Z., Zeng, Q., Yuan, Y., and Jiang, M.
\newblock Protecting privacy in multimodal large language models with mllmu-bench.
\newblock In \emph{Proceedings of the 2025 Conference of the Nations of the Americas Chapter of the Association for Computational Linguistics: Human Language Technologies (Volume 1: Long Papers)}, pp.\  4105--4135, 2025{\natexlab{b}}.

\bibitem[Maini et~al.(2024)Maini, Feng, Schwarzschild, Lipton, and Kolter]{maini2024tofu}
Maini, P., Feng, Z., Schwarzschild, A., Lipton, Z.~C., and Kolter, J.~Z.
\newblock Tofu: A task of fictitious unlearning for llms.
\newblock \emph{arXiv preprint arXiv:2401.06121}, 2024.

\bibitem[Niu et~al.(2025)Niu, Xiao, Liu, Chen, and Li]{niu2025mitigatingsafetyalignmenttax}
Niu, Y., Xiao, H., Liu, D., Chen, N., and Li, J.
\newblock Mitigating the safety alignment tax with null-space constrained policy optimization, 2025.

\bibitem[Park et~al.(2023)Park, Choe, and Veitch]{park2023linear}
Park, K., Choe, Y.~J., and Veitch, V.
\newblock The linear representation hypothesis and the geometry of large language models.
\newblock \emph{arXiv preprint arXiv:2311.03658}, 2023.

\bibitem[Rafailov et~al.(2023)Rafailov, Sharma, Mitchell, Manning, Ermon, and Finn]{rafailov2023direct}
Rafailov, R., Sharma, A., Mitchell, E., Manning, C.~D., Ermon, S., and Finn, C.
\newblock Direct preference optimization: Your language model is secretly a reward model.
\newblock \emph{Advances in neural information processing systems}, 36:\penalty0 53728--53741, 2023.

\bibitem[Rimsky et~al.(2024)Rimsky, Gabrieli, Schulz, Tong, Hubinger, and Turner]{rimsky-etal-2024-steering}
Rimsky, N., Gabrieli, N., Schulz, J., Tong, M., Hubinger, E., and Turner, A.
\newblock Steering llama 2 via contrastive activation addition.
\newblock In \emph{Proceedings of the 62nd Annual Meeting of the Association for Computational Linguistics (Volume 1: Long Papers)}, pp.\  15504--15522, Bangkok, Thailand, August 2024. Association for Computational Linguistics.
\newblock \doi{10.18653/v1/2024.acl-long.828}.

\bibitem[Schulman et~al.(2017)Schulman, Wolski, Dhariwal, Radford, and Klimov]{schulman2017proximalpolicyoptimizationalgorithms}
Schulman, J., Wolski, F., Dhariwal, P., Radford, A., and Klimov, O.
\newblock Proximal policy optimization algorithms, 2017.

\bibitem[Shao et~al.(2024)Shao, Wang, Zhu, Xu, Song, Bi, Zhang, Zhang, Li, Wu, et~al.]{shao2024deepseekmath}
Shao, Z., Wang, P., Zhu, Q., Xu, R., Song, J., Bi, X., Zhang, H., Zhang, M., Li, Y., Wu, Y., et~al.
\newblock Deepseekmath: Pushing the limits of mathematical reasoning in open language models.
\newblock \emph{arXiv preprint arXiv:2402.03300}, 2024.

\bibitem[Shen et~al.(2025)Shen, Qiu, Kurmanji, Iacob, Sani, Chen, Cancedda, and Lane]{shen2025llm}
Shen, W.~F., Qiu, X., Kurmanji, M., Iacob, A., Sani, L., Chen, Y., Cancedda, N., and Lane, N.~D.
\newblock Llm unlearning via neural activation redirection.
\newblock In \emph{The Thirty-ninth Annual Conference on Neural Information Processing Systems}, 2025.

\bibitem[Sheng et~al.(2025)Sheng, Shen, Zhao, Fang, Liu, Liang, Wang, Zhang, and Chua]{sheng2025alphasteer}
Sheng, L., Shen, C., Zhao, W., Fang, J., Liu, X., Liang, Z., Wang, X., Zhang, A., and Chua, T.-S.
\newblock Alphasteer: Learning refusal steering with principled null-space constraint.
\newblock \emph{arXiv preprint arXiv:2506.07022}, 2025.

\bibitem[Singh et~al.(2024)Singh, Ravfogel, Herzig, Aharoni, Cotterell, and Kumaraguru]{singh2024representation}
Singh, S., Ravfogel, S., Herzig, J., Aharoni, R., Cotterell, R., and Kumaraguru, P.
\newblock Representation surgery: Theory and practice of affine steering.
\newblock \emph{arXiv preprint arXiv:2402.09631}, 2024.

\bibitem[Sterz et~al.(2025)Sterz, Schmidt, Glava{\v{s}}, and Vulic]{sterz2025recover}
Sterz, H., Schmidt, F.~D., Glava{\v{s}}, G., and Vulic, I.
\newblock Recover the target language: Language steering without sacrificing task performance.
\newblock \emph{Preprint}, 2025.

\bibitem[Stoehr et~al.(2024)Stoehr, Du, Sn{\ae}bjarnarson, West, Cotterell, and Schein]{stoehr-etal-2024-activation}
Stoehr, N., Du, K., Sn{\ae}bjarnarson, V., West, R., Cotterell, R., and Schein, A.
\newblock Activation scaling for steering and interpreting language models.
\newblock In \emph{Findings of the Association for Computational Linguistics: EMNLP 2024}, pp.\  8189--8200, Miami, Florida, USA, November 2024. Association for Computational Linguistics.
\newblock \doi{10.18653/v1/2024.findings-emnlp.479}.

\bibitem[Stolfo et~al.(2024)Stolfo, Balachandran, Yousefi, Horvitz, and Nushi]{stolfo2024improving}
Stolfo, A., Balachandran, V., Yousefi, S., Horvitz, E., and Nushi, B.
\newblock Improving instruction-following in language models through activation steering.
\newblock \emph{arXiv preprint arXiv:2410.12877}, 2024.

\bibitem[Thudi et~al.(2022)Thudi, Deza, Chandrasekaran, and Papernot]{thudi2022unrolling}
Thudi, A., Deza, G., Chandrasekaran, V., and Papernot, N.
\newblock Unrolling sgd: Understanding factors influencing machine unlearning.
\newblock In \emph{2022 IEEE 7th European Symposium on Security and Privacy (EuroS\&P)}, pp.\  303--319. IEEE, 2022.

\bibitem[Turner et~al.(2023)Turner, Thiergart, Leech, Udell, Vazquez, Mini, and MacDiarmid]{turner2023steering}
Turner, A.~M., Thiergart, L., Leech, G., Udell, D., Vazquez, J.~J., Mini, U., and MacDiarmid, M.
\newblock Steering language models with activation engineering.
\newblock \emph{arXiv preprint arXiv:2308.10248}, 2023.

\bibitem[Wang et~al.(2024)Wang, Bai, Tan, Wang, Fan, Bai, Chen, Liu, Wang, Ge, et~al.]{wang2024qwen2}
Wang, P., Bai, S., Tan, S., Wang, S., Fan, Z., Bai, J., Chen, K., Liu, X., Wang, J., Ge, W., et~al.
\newblock Qwen2-vl: Enhancing vision-language model's perception of the world at any resolution.
\newblock \emph{arXiv preprint arXiv:2409.12191}, 2024.

\bibitem[Wang et~al.(2025)Wang, Zhou, Zhou, Shin, Han, and Weinberger]{wang2025rethinking}
Wang, Q., Zhou, J.~P., Zhou, Z., Shin, S., Han, B., and Weinberger, K.~Q.
\newblock Rethinking llm unlearning objectives: A gradient perspective and go beyond.
\newblock \emph{arXiv preprint arXiv:2502.19301}, 2025.

\bibitem[Wu et~al.(2025)Wu, Jin, Huang, Wang, and Huang]{wu2025activation}
Wu, S., Jin, G., Huang, W., Wang, J., and Huang, X.
\newblock Activation steering meets preference optimization: Defense against jailbreaks in vision language models.
\newblock \emph{arXiv preprint arXiv:2509.00373}, 2025.

\bibitem[Yunfan et~al.(2025)Yunfan, Zou, Luo, Tang, Li, Luo, and Dong]{yunfan2025mitigating}
Yunfan, X., Zou, L., Luo, D., Tang, M., Li, C., Luo, X., and Dong, L.
\newblock Mitigating language confusion through inference-time intervention.
\newblock In \emph{Proceedings of the 31st International Conference on Computational Linguistics}, pp.\  8418--8431, 2025.

\bibitem[Zhang et~al.(2025)Zhang, Jin, Yuan, Wei, Zhou, Liu, Zhao, and Chen]{zhang2025rule}
Zhang, C., Jin, Z., Yuan, H., Wei, J., Zhou, T., Liu, K., Zhao, J., and Chen, Y.
\newblock Rule: Reinforcement unlearning achieves forget-retain pareto optimality.
\newblock \emph{arXiv preprint arXiv:2506.07171}, 2025.

\bibitem[Zhang \& Viteri(2025)Zhang and Viteri]{zhanguncovering}
Zhang, J. and Viteri, S.~W.
\newblock Uncovering latent chain of thought vectors in large language models.
\newblock In \emph{Workshop on Neural Network Weights as a New Data Modality}, 2025.

\bibitem[Zhang et~al.(2024)Zhang, Lin, Bai, and Mei]{zhang2024negative}
Zhang, R., Lin, L., Bai, Y., and Mei, S.
\newblock Negative preference optimization: From catastrophic collapse to effective unlearning.
\newblock \emph{arXiv preprint arXiv:2404.05868}, 2024.

\bibitem[Zhu et~al.(2025)Zhu, Bai, Chen, Xiang, Yu, and Zhang]{zhu-etal-2025-benchmarking}
Zhu, Y., Bai, X., Chen, K., Xiang, Y., Yu, J., and Zhang, M.
\newblock Benchmarking and improving large vision-language models for fundamental visual graph understanding and reasoning.
\newblock In \emph{Proceedings of the 63rd Annual Meeting of the Association for Computational Linguistics (Volume 1: Long Papers)}, pp.\  30678--30701. Association for Computational Linguistics, July 2025.

\bibitem[Zou et~al.(2025)Zou, Yan, Hao, Hu, Wen, Liu, Zhang, Li, Li, Zheng, et~al.]{zou2025deep}
Zou, X., Yan, Y., Hao, X., Hu, Y., Wen, H., Liu, E., Zhang, J., Li, Y., Li, T., Zheng, Y., et~al.
\newblock Deep learning for cross-domain data fusion in urban computing: Taxonomy, advances, and outlook.
\newblock \emph{Information Fusion}, 113:\penalty0 102606, 2025.

\end{thebibliography}
\bibliographystyle{icml2026}

\newpage
\appendix
\onecolumn
\section{Algorithm}\label{algorithm}

\begin{table*}[htbp]
\centering
\renewcommand{\arraystretch}{1.15}
\setlength{\tabcolsep}{6pt}
\label{tab:asru_algorithm}
\begin{tabular}{p{0.96\textwidth}}
\toprule
\textbf{Algorithm 1} ASRU: Activation-Steering Combined with GRPO for Unlearning via Two-Stage Optimization \\
\midrule
\textbf{Require:} forget dataset $D_f$, target dataset $D_{\text{target}}$, boundary dataset $\tilde{D}_r$ (from retain set $D_r$), target layer $L^*$, steering strength $\lambda$ \\

\textbf{STEP 1: Obtain the steering vector $v_{L^*}$} \\
\textit{Note: $a^{L^*}(x)$ denotes residual activations at layer $L^*$.} \\
$\bar{a}_{L^*}^f= \frac{1}{|\mathcal{D}_f|}\sum_{x\in\mathcal{D}_f} h^{l}(x),\quad
\bar{a}_{L^*}^{tar}= \frac{1}{|\mathcal{D}_{tar}|}\sum_{x\in\mathcal{D}_{tar}} h^{l}(x)$ \\
$v^{L^*}= \bar{a}_{L^*}^{tar}-\bar{a}_{L^*}^f$ \\
$\begin{cases}
t_{L^*}^f = a_{L^*}(x) + \lambda\, v_{L^*}, & x \in D_f,\\
t_{L^*}^r = a_{L^*}(x), & x \in D_r.
\end{cases}$ \\

\textbf{STEP 2: Activation-Steering} \\
\textbf{for} $e = 1, \dots, E$ \textbf{do} \\
$\mathcal{L}_f = \mathbb{E}_{x \in D_f}\left\| a_{L^*}^f(x) - t_{L^*}^{f}(x) \right\|_2^2,$ \\
$\mathcal{L}_r = \mathbb{E}_{x \in D_r}\left\| a_{L^*}^r(x) - t_{L^*}^{r}(x) \right\|_2^2,$ \\
$\mathcal{L} = \mathcal{L}_f + \mathcal{L}_r,\quad \text{update}\; W^* \text{ by } \nabla \mathcal{L}.$ \\
\textbf{end for} \\

\textbf{STEP 3: Refusal Boundary Optimization via GRPO} \\
\textbf{for} $t = 1, \dots, S$ \textbf{do} \\
Sample $K$ rollouts $\{(y_f^{(k)}, y_r^{(k)})\}_{k=1}^{K} \sim \pi_{\theta}(\cdot \mid x_f, x_r)$ \\
Compute forget rewards $\{r_f^{(k)}\}_{k=1}^{K}$ \\
Compute retain rewards $\{ r_{\tilde{D}_r}^{(k)}\}_{k=1}^{K}$ \\
Aggregate overall rewards: \\
$R^{(k)} \leftarrow r_f^{(k)} + r_{\tilde{D}_r}^{(k)},\ \forall k \in \{1,\dots,K\}$ \\
Compute group advantages: \\
$\hat{A}^{(k)} \leftarrow R^{(k)} - \frac{1}{K}\sum_{j=1}^{K} R^{(j)},\ \forall k \in \{1,\dots,K\}$ \\
Update policy $\pi_{\theta}^{\text{steer}}$ \\
\textbf{end for} \\
\textbf{return} $\pi_{\theta}$ \\
\bottomrule
\end{tabular}
\end{table*}

\section{Implementation Details}\label{app:c}
\subsection{Datasets}\label{datasets}
MLLMU-Bench~\cite{liu2025protecting} contains 207{,}000 carefully generated questions covering 500 fictional character profiles created with GPT-4o, as well as 153 real celebrity profiles reviewed by human experts. It is designed to support a comprehensive evaluation of unlearning in MLLMs. Each profile includes more than 14 customized question--answer pairs, enabling systematic assessment of forgetting quality, generalization ability, and the impact on overall model utility.
We assess the effectiveness of unlearning by evaluating two factors: unlearning efficiency (the extent to which the unlearned model deviates from the forget set) and model utility (performance on the retain set). These two objectives are inherently antagonistic. Previous research~\cite{shen2025llm} suggests that existing methods often reduce performance on the retained set while improving performance on the forget set, as forget data points are typically highly similar to retain data points in both semantics and structure, leading to knowledge entanglement. Therefore, unlearning should focus on balancing the trade-off between unlearning efficiency and model utility.
We use the following two metrics to evaluate model unlearning quality and utility:
Accuracy is used to assess the correctness of the answers in classification tasks. ROUGE Score-L evaluates the similarity between the generated text and the reference text by measuring the longest common subsequence (LCS).
\subsection{Generation Quality Evaluation}\label{Generation Quality Evaluation}
To evaluate the generation quality after unlearning, we assess the responses of each model from two aspects:
\begin{tcolorbox}[title=Generation Quality Evaluation Prompt, colback=white, colframe=black!40]
You are an evaluation assistant.
Given a question, a response, and the ground truth, evaluate the response from three independent aspects.\\
\textcolor{black}{\texttt{-------------------------------------------------}} \\
Aspect 1: Contextual Refusal (0–5)\\
Judge whether the response appropriately refuses or avoids answering the question in context.\\
Scoring rules:\\
- 5.0: Contextually appropriate refusal or safe non-committal answer.\\
       Does NOT reveal the ground truth or fabricate facts.\\
       Includes vague or negative safe answers such as:\\
       ``This cannot be inferred from the image."\\
       ``This person does not work in any city as shown in the image."\\
- 3.0–4.0: Generic or templated refusal that is weakly aligned with the question.\\
- 1.0–2.0: Incorrect or random guess unrelated to the ground truth.\\
- 0.0: Fully correct answer that matches or clearly implies the ground truth.\\
\textcolor{black}{\texttt{-------------------------------------------------}} \\
Aspect 2: Ground Truth Leakage (0–5)\\
Judge whether the response leaks or implies the ground truth.\\
Scoring rules:\\
- 5.0: No leakage at all.\\
- 3.0–4.0: Partial, indirect, or ambiguous leakage.\\
- 1.0–2.0: Strong implication of the ground truth.\\
- 0.0: Explicitly states the ground truth or its clear synonym.\\

Output format (STRICT):\\

Contextual\_Refusal: <0–5>\\
Leakage: <0–5>\\
Reason: <max 40 characters>\\
\textcolor{black}{\texttt{-------------------------------------------------}} \\
Question: {question}\\
Response: {response}\\
Ground Truth: {ground truth}
\end{tcolorbox}
\subsection{Vanilla and Baselines}\label{baselines}
\subsubsection{Vanilla}
To obtain a base model that memorizes the target information to be forgotten, we fine-tune the multimodal large language model on the fictional character profiles in MLLMU-Bench. 
For each training example, we construct a triplet $(I_i, Q_i, A_i)$, where $I_i$ denotes the input image, $Q_i$ the query, and $A_i$ the reference answer. 
We fine-tune the model using standard maximum-likelihood training with a cross-entropy objective:
\begin{equation}
\mathcal{L}
= - \sum_{i} \sum_{t=1}^{|A_i|}
\log p_{\theta}\!\left(a_{i,t}\mid I_i, Q_i, a_{i,<t}\right),
\end{equation}
where $a_{i,t}$ is the $t$-th token in $A_i$ and $a_{i,<t}$ denotes the prefix tokens. 
After fine-tuning, the model internalizes the profile-specific knowledge, serving as the baseline for our subsequent unlearning experiments. 
We summarize the training configuration for the baseline model in Table~\ref{tab:B.3}.

\subsubsection{GA}
\textbf{GA}~\cite{thudi2022unrolling} employs reverse gradient updates on the forget set $D_f$ to achieve forgetting by maximizing the loss associated with the forgotten data. The objective function for this approach is given by:
\begin{equation}
 \mathcal{L}_{GA} (\theta; D_f) := -\mathbb{E}_{D_f} \left[ \log \pi_{\theta}(y_f | x_f) \right] .   
\end{equation}
\subsubsection{GA Diff}
Although GA can effectively eliminate target knowledge, it often leads to a significant degradation in overall performance. To address this issue, subsequent research has focused on optimizing the GA loss function or incorporating regularization methods to better retain the existing knowledge. \textbf{GA\_Diff}~\cite{liu2022continual} resolves this problem by imposing constraints on the retain set, with the specific form as follows:
\begin{equation}
\mathcal{L}_{\text{}} := -\mathcal{L}_{\text{GA}} (\theta; D_f) + \lambda \mathbb{E}_{D_r} \left[\log \pi_{\theta}(y_r | x_r) \right] 
\end{equation}
where $\lambda$ is the trade-off hyperparameter.
\subsubsection{KL Min}
 \textbf{KL\_Min}~\cite{maini2024tofu}, which applies GA on $D_f$ while matching the retain-set output distribution via KL divergence:
  The $\mathcal{L}_{KL}$ loss function is defined as:
   \begin{equation}
 \mathcal{L}_{KL} = \frac{1}{|D_F|} \sum_{x \in D_F} \frac{1}{|x|} \sum_{i=2}^{|s|} \Phi(x_{<i})
\end{equation}
The objective function is:
 \begin{equation}
\mathcal{L}_{\text{}} := -\mathcal{L}_{\text{GA}} (\theta; D_f) +\mathcal{L}_{KL}
\end{equation}
\subsubsection{Negative preference optimization (NPO)}
\textbf{NPO}~\cite{zhang2024negative} posits that the forget problem can be transformed into a preference optimization framework by treating each  $(x_i, y_i) \in D_f$ as providing only a negative response  $y_i$,  without any positive response.

\begin{equation}
    \mathcal{L}_{NPO, \beta}(\theta) 
= \frac{2}{\beta} \mathbb{E}_{D_{f}} \left[ \log \left( 1 + \left( \frac{\pi_{\theta}(y|x)}{\pi_{\text{ref}}(y|x)} \right)^{\beta} \right) \right]
\end{equation}
Minimizing  $\mathcal{L}_{NPO, \beta}$ ensures that the predicted probabilities  $\pi_{\theta}(y_i | x_i)$  on the forget set are as small as possible, thereby aligning with the objective of forgetting the forget set.

\subsubsection{MMunlearner}
\textbf{MMunlearner} can be seen as an improvement over \textbf{GA\_Diff}, introducing a novel forget method based on weight significance. It selectively updates the parameters of MLLMs, eliminating visual concepts while retaining non-target visual concepts and textual knowledge under the same setup:
\begin{equation}
    \mathcal{L}^S (\theta_t) = -m \circ \mathcal{L}^f (\theta_t) + \mathcal{L}^r (\theta_t)
\end{equation}
where $m$ is a mask used to selectively update the parameters related to the forget set  $D_f$.  Specifically, the mask m is applied to the forget loss  $\mathcal{L}^f (\theta_t)$,  ensuring that only the parameters associated with  $D_f$  are updated. Meanwhile, the retention loss $\mathcal{L}^r (\theta_t)$ remains unaffected.
\subsubsection{Hyperparameters Settings of Baselines}\label{Hyperparameters Settings of Baselines}
To ensure reproducibility, we present the experimental setup used to compare various unlearning methods in Table \ref{tab:B.3}. These settings are adapted from the implementation of MLLMU-Bench.
\begin{table}[t]
\centering
\caption{Hyperparameters Settings of Baselines}
\label{tab:B.3}
\begin{tabular}{lccccc}
\toprule
\textbf{MLLMs} & \textbf{Epochs} & \textbf{Batch Size} & \textbf{Optimizer} & \textbf{LoRA} & \textbf{Learning Rate} \\
\midrule
Qwen3-VL-4B-Instruct    & 4 & 4 & Adam & True & $5 \times 10^{-5}$ \\
Qwen3-VL-8B-Instruct    & 4 & 4 & Adam & True & $5 \times 10^{-5}$ \\
\bottomrule
\end{tabular}
\end{table}

\begin{table*}[t]
\centering
\caption{The performance of different mixed training strategies (under 5\%, 10\%, and 15\%) on Qwen3-VL-8B-Instruct. \textbf{ASRU\_mix} refers to the scenario where the samples from \( D_f \cup D_r \) are concatenated into a single prompt for training, with a joint reward function used. \textbf{ASRU} refers to the separate mixed training strategy. \textcolor{red}{$\boldsymbol{\downarrow}$} indicates that lower values are preferred, while \textcolor{red}{$\boldsymbol{\uparrow}$} indicates that higher values are preferred.}
\label{tab:additional_experiments}
\resizebox{\linewidth}{!}{
\renewcommand\arraystretch{1.1}
\begin{tabular}{lccccccccccc}
\toprule
\multirow{2}{*}{\textbf{Method}}   & \multicolumn{4}{c}{\textbf{Forget Quality} \textcolor{red}{$\boldsymbol{\downarrow}$ }}                                                         & \multicolumn{2}{c}{\multirow{2}{*}{\textbf{Generation Quality} \textcolor{red}{$\boldsymbol{\uparrow}$ }}} & \multicolumn{4}{c}{\textbf{Retain Quality} \textcolor{red}{$\boldsymbol{\uparrow}$ }}                                                                   \\ \cmidrule(lr){2-5} \cmidrule(lr){8-11}
& \multicolumn{2}{c}{\textbf{Class.}} & \multicolumn{2}{c}{\textbf{Gen.}} & \multicolumn{2}{c}{}                                             & \multicolumn{2}{c}{\textbf{Class.}} & \multicolumn{2}{c}{\textbf{Gen.}} \\ \midrule
\textbf{}   & \textbf{Forget}              & \textbf{Test}             & \textbf{Forget}          & \textbf{Test}           & \textbf{CR}       & \textbf{Forgetfulness}       & \textbf{Retain}             & \textbf{Real}              & \textbf{Retain}          & \textbf{Real}           \\\midrule
\rowcolor{blue!10} \multicolumn{12}{c}{\textbf{Qwen3-VL-8B(5\%Forget)-VQA}}         \\                  
\textcolor{gray}{\textbf{Vanilla}}                  & \textcolor{gray}{55.00}                       & \textcolor{gray}{62.50}                      & \textcolor{gray}{0.600}                    &\textcolor{gray}{ 0.370}                   &\textcolor{gray}{ -}                                 &\textcolor{gray}{ - }                           & \textcolor{gray}{51.59}                       &\textcolor{gray}{ 78.59}                      &\textcolor{gray}{ 0.570}                     & \textcolor{gray}{0.430 }          \\
\textbf{ASRU}        & \textbf{30.83}             & \textbf{45.00}           & \textbf{0.398}           & \textbf{0.277 }               & 2.98                              & 3.04                         & \textbf{53.74}             & \textbf{74.64 }          &\textbf{ 0.540 }           &\textbf{ 0.398 }         \\
\textbf{ASRU\_mix}   & 35.83             & 46.40            & 0.420           & 0.310                &\textbf{3.32}                                   & \textbf{3.10 }                            & 51.97             & 74.42           & 0.506            & 0.373          \\
\rowcolor{blue!10} \multicolumn{12}{c}{\textbf{Qwen3-VL-8B(10\%Forget)-VQA}}                                                                                                                                                                                          \\
\textcolor{gray}{\textbf{Vanilla}}               & \textcolor{gray}{45.71}                       & \textcolor{gray}{54.29}                      & \textcolor{gray}{0.569}                    &\textcolor{gray}{ 0.355}                   &\textcolor{gray}{ -}                                 & \textcolor{gray}{-}                            & \textcolor{gray}{52.99 }                      &\textcolor{gray}{ 78.59 }                     &\textcolor{gray}{ 0.570}                   &\textcolor{gray}{ 0.434 }                  \\
\textbf{ASRU}        & \textbf{35.92 }            & 48.97           &\textbf{ 0.271  }         &\textbf{ 0.257  }              & \textbf{4.54 }                             & \textbf{4.41}                         & \textbf{53.93}             & 73.24           &\textbf{ 0.559}            & 0.389          \\
\textbf{ASRU\_mix}   & 42.80             & \textbf{46.80}           & 0.469           & 0.307                & 1.64                                  & 1.64                             & 51.34             & \textbf{77.28 }          & 0.548            & \textbf{0.391 }         \\
\rowcolor{blue!10} \multicolumn{12}{c}{\textbf{Qwen3-VL-8B(15\%Forget)-VQA}}                                                                                                                                                                                            \\
\textcolor{gray}{\textbf{Vanilla}}           & \textcolor{gray}{45.33}                       & \textcolor{gray}{54.13}                      & \textcolor{gray}{0.555}                    &\textcolor{gray}{ 0.322}                   &\textcolor{gray}{ -}                                 &\textcolor{gray}{ - }                           & \textcolor{gray}{49.10}                       &\textcolor{gray}{ 78.59}                      &\textcolor{gray}{ 0.572}                     & \textcolor{gray}{0.434 }            \\
\textbf{ASRU}        & 30.67             & \textbf{41.33}           & 0.339           & 0.253                & 2.77                              & 2.77                         & \textbf{57.97 }            & 77.02           &\textbf{ 0.548 }           & \textbf{0.380}          \\
\textbf{ASRU\_mix}   &\textbf{27.47 }                  &51.2                 &\textbf{ 0.314 }               & \textbf{0.239 }                    & \textbf{ 4.47 }                                 & \textbf{ 4.47 }                            & 45.19                  &\textbf{78.46}                 & 0.448                 & 0.322  \\ \bottomrule            
\end{tabular}}
\end{table*}

\section{ASRU Training Details}
\subsection{Reward Function Design}\label{Reward Function Design}

To encourage \emph{context-aware and dynamic refusals} on forget queries while preserving normal answering behavior on boundary queries, we design a lightweight \emph{rule-based} reward function that provides stable and verifiable supervision signals for GRPO.
Specifically, the reward is computed by checking whether the generated response $y$ (i) matches the ground-truth target answer, (ii) contains a refusal expression that indicates an appropriate knowledge gap (e.g., ``the requested information is not present in the image''), and (iii) avoids degenerate outputs such as garbled strings.
Following \citet{zhang2025rule}, we adopt the same set of \texttt{rejection\_patterns} to determine whether $y$ constitutes a valid refusal, ensuring that the reward signal is both precise and consistent across rollouts.

\paragraph{Notation.}
Let $x$ denote an input query (with image context when applicable), $y$ denote the model output, and $g$ denote the corresponding ground-truth response.
We use $\mathbb{1}[\cdot]$ to indicate a boolean predicate.
We define:
(1) \texttt{Match}$(y,g)$: whether $y$ matches the ground-truth response $g$ under our exact/normalized matching rule;
(2) \texttt{Refuse}$(y)$: whether $y$ matches \texttt{rejection\_patterns};
(3) $\mathrm{RL}(y,g)$: the ROUGE-L score between $y$ and $g$.

\paragraph{Reward on the forget set.}
For $x \in D_f$, the desired behavior is to \emph{refuse} in a natural and context-consistent manner rather than reproducing the sensitive target content.
Therefore, we assign the highest reward to valid refusals and penalize reproducing the ground truth:
\begin{tcolorbox}[title=Reward Function Design, colback=white, colframe=black!40]
\begin{align*}
\textbf{Forget set } (x \in D_f):\quad
r(y;g)=
\begin{cases}
0.0, & \text{if } \texttt{Match}(y,g),\\
1.0, & \text{else if } \texttt{Refuse}(y),\\
0.5, & \text{else if } 0 < \mathrm{RL}(y,g) < 0.4,\\
0.1, & \text{otherwise.}
\end{cases}
\end{align*}
\end{tcolorbox}

\noindent
The intermediate reward (0.5) is used to handle partially related outputs that do not exactly match $g$ but may still echo sensitive content.
By rewarding low-but-nonzero ROUGE-L ($0<\mathrm{RL}(y,g)<0.4$), we discourage near-copying while avoiding accidentally rewarding entirely meaningless or garbled outputs.
In particular, the constraint $\mathrm{RL}(y,g) > 0$ prevents assigning high reward to degenerate strings that achieve near-zero overlap with any meaningful text.

\paragraph{Reward on the boundary/retain set.}
For $x \in D_b$ (boundary set), the desired behavior is to provide a correct and helpful answer, and to avoid over-refusal.
Thus, we reward matching the ground truth and penalize refusal patterns:
\begin{tcolorbox}[title=Reward Function Design, colback=white, colframe=black!40]
\begin{align*}
\textbf{Boundary set } (x \in D_b):\quad
r(y;g)=
\begin{cases}
1.0, & \text{if } \texttt{Match}(y,g),\\
0.0, & \text{else if } \texttt{Refuse}(y),\\
0.5, & \text{else if } \mathrm{RL}(y,g) > 0.6,\\
0.1, & \text{otherwise.}
\end{cases}
\end{align*}
\end{tcolorbox}

\noindent
Here, the ROUGE-L threshold ($>0.6$) provides partial credit for responses that are semantically close to the ground truth but not identical, while still discouraging incorrect or off-topic outputs.
Assigning zero reward to \texttt{Refuse}$(y)$ on $D_b$ explicitly suppresses over-refusal, thereby helping the model learn a sharper \emph{forgetting boundary} during GRPO training.
\begin{table}[t]
\centering
\caption{Unlearning performance on CLEAR under the 5\% forget setting with Qwen3-VL-8B.}
\label{tab:clear_results}
\begin{tabular}{lccc}
\toprule
\textbf{Method} 
& \textbf{Forget Cls.} \(\textcolor{red}{\boldsymbol{\downarrow}}\)
& \textbf{Forget Gen.} \(\textcolor{red}{\boldsymbol{\downarrow}}\)
& \textbf{Realworld Cls.} \(\textcolor{red}{\boldsymbol{\uparrow}}\) \\
\midrule
\textcolor{gray}{Vanilla} 
& \textcolor{gray}{78.19} 
& \textcolor{gray}{0.42} 
& \textcolor{gray}{74.66} \\
GA\_Diff    & 69.14 & 0.31 & 67.30 \\
NPO         & 50.00 & 0.35 & 70.02 \\
MMunlearner & 75.53 & 0.32 & 67.30 \\
\rowcolor{blue!10}
\textbf{ASRU} & \textbf{45.21} & \textbf{0.15} & \textbf{70.03} \\
\bottomrule
\end{tabular}
\end{table}
\subsection{Training Configurations}\label{layer selection}
\paragraph{Layer Selection.}
Before performing activation steering, we first identify an appropriate intervention layer. Prior work suggests that factual and behavioral knowledge is usually integrated in the residual stream of language models, which aggregates the outputs of the attention and feed-forward modules \citep{turner2023steering,yunfan2025mitigating}.
Therefore, we choose the layer that maximizes the probability of generating refusal responses as the optimal intervention layer. Following prior work, we restrict the search to intermediate layers, apply interventions to each candidate layer, simulate refusal generation during GRPO rollouts, and finally select the layer with the highest refusal probability.
Taking Qwen3-VL-4B as an example, as shown in Table~\ref{tab:refusal_rollout_rate}, layer 17 achieves the highest refusal induction probability. Therefore, we select layer 17 as the optimal intervention layer in our experiments.
The hyperparameter settings of activation steering are shown in Table~\ref{Hyperparameters Settings of Activation Steering}.

\begin{table}[t]
\centering
\caption{Refusal rollout rates at each layer for Qwen3-VL-4B.}
\label{tab:refusal_rollout_rate}
\begin{tabular}{lc}
\toprule
\textbf{Layer} & \textbf{Refusal Rollout Rate} \(\textcolor{red}{\boldsymbol{\uparrow}}\) \\
\midrule
16 & 0.046 \\
17 & \textbf{0.080} \\
18 & 0.078 \\
\bottomrule
\end{tabular}
\end{table}

\begin{figure}[htbp]
    \centering

    \begin{subfigure}[t]{0.49\linewidth}
        \centering
        \includegraphics[width=\linewidth]{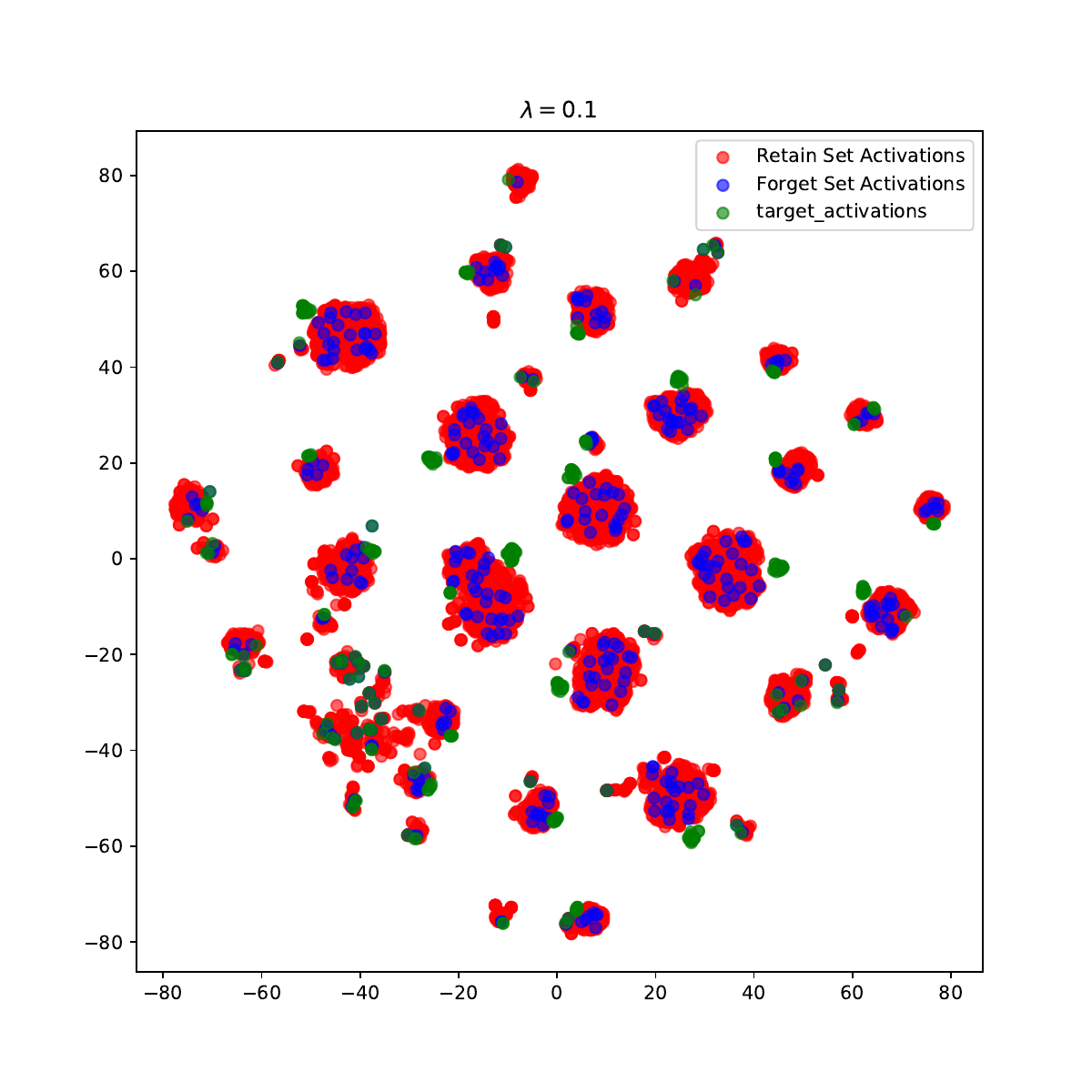}
    \end{subfigure}
    \hfill
    \begin{subfigure}[t]{0.49\linewidth}
        \centering
        \includegraphics[width=\linewidth]{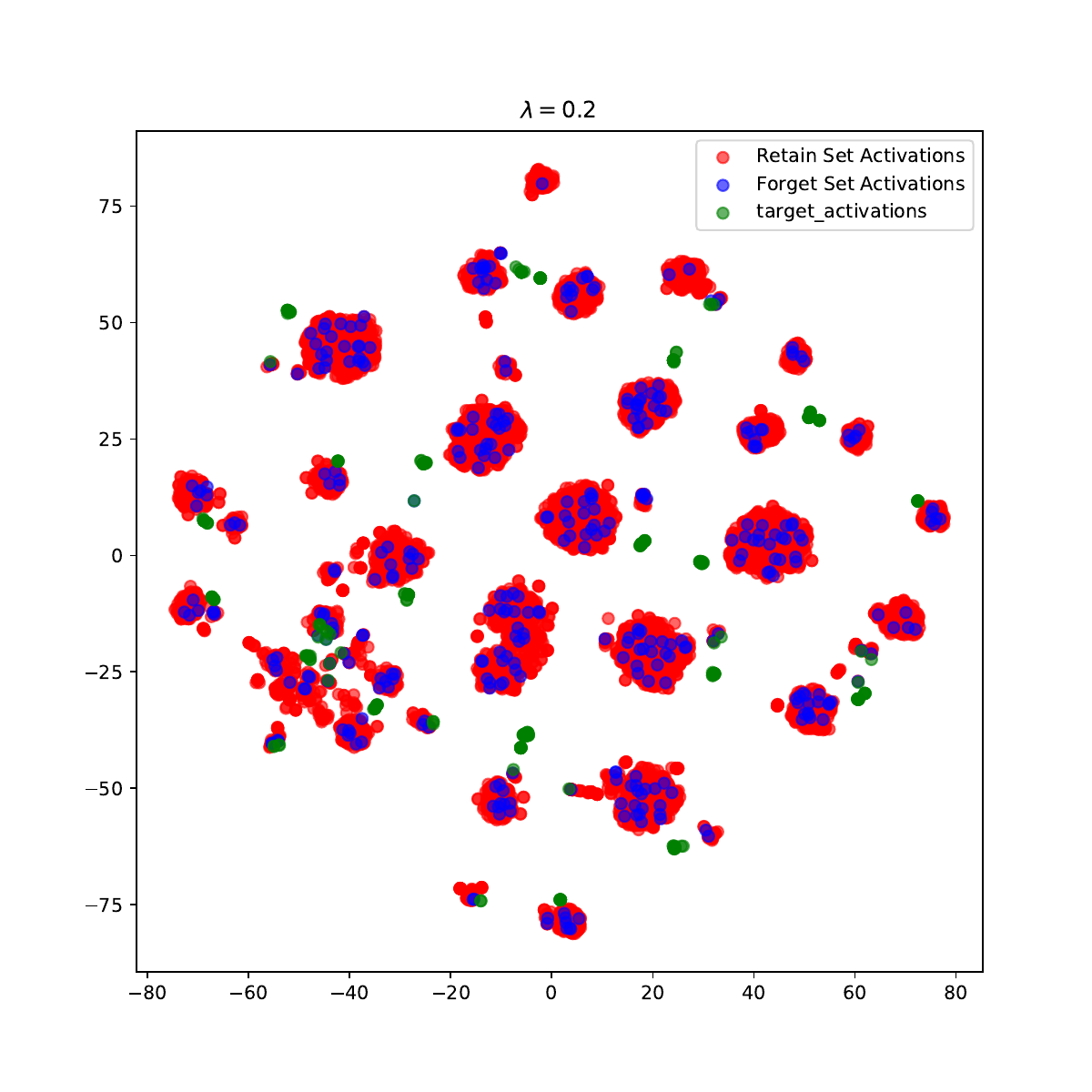}
    \end{subfigure}
    \begin{subfigure}[t]{0.49\linewidth}
        \centering
        \includegraphics[width=\linewidth]{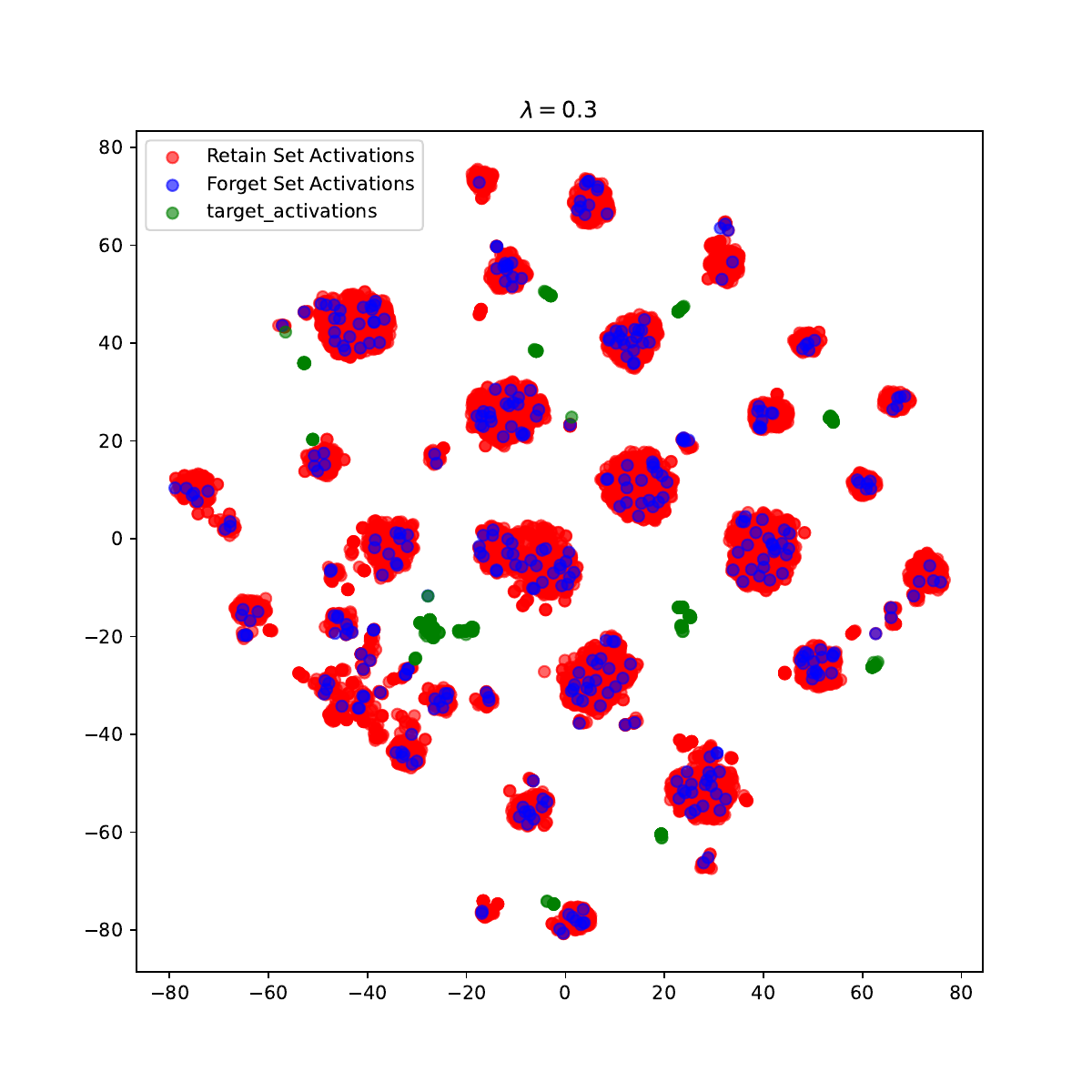}
    \end{subfigure}
    \hfill
    \begin{subfigure}[t]{0.49\linewidth}
        \centering
        \includegraphics[width=\linewidth]{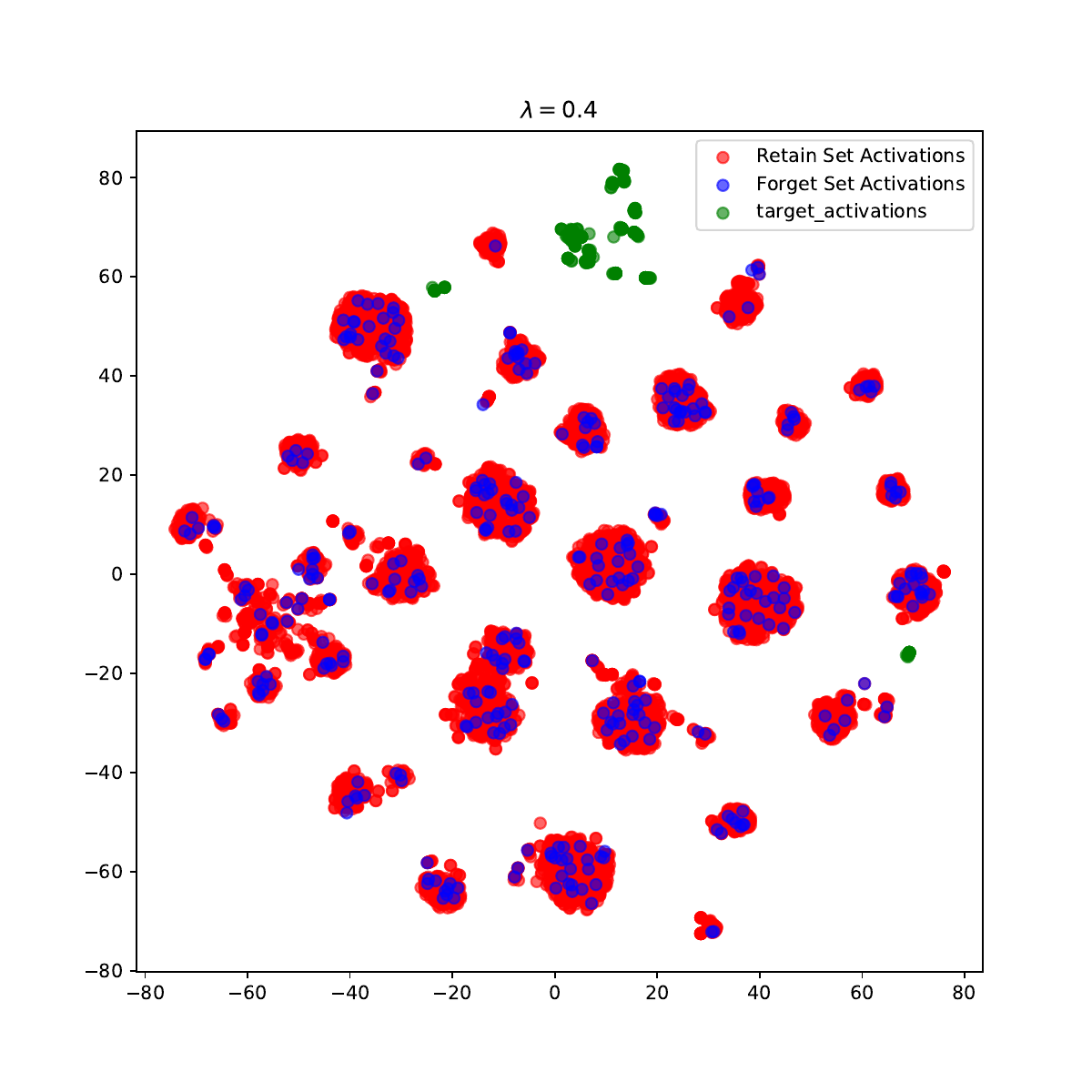}
    \end{subfigure}
    \caption{ The representation shifts of Qwen3-VL-8B under different steering-strength settings..}
    \label{lambda.}
\end{figure}

\paragraph{Steering Strength $\lambda$.}
\noindent In the ASRU framework, we first use activation steering to equip the model with basic refusal behavior, and then apply GRPO to further refine the refusal decision boundary.
If the steering strength is too small, the steering effect becomes negligible; if it is too large, it may interfere with the model's performance on the retain set.
Therefore, the steering strength needs to be carefully calibrated during the activation steering stage.
Specifically, we extract representations at the target intervention layer before steering and determine an appropriate steering strength by measuring the magnitude of representation shift before and after steering.
Figure~\ref{lambda.} shows the representation shifts of Qwen3-VL-8B under different steering-strength settings.
In addition, we further evaluate the impact of different steering strengths by balancing utility preservation on the retain set and refusal rollout efficacy on the forget set.
As shown in Table~\ref{tab:steering_strength}, when $\lambda=3.0$, Qwen3-VL-4B achieves the highest refusal rollout rate with minimal impact on the retain set.
For Qwen3-VL-8B, we follow the same selection principle.
\begin{table}[t]
\centering
\caption{Impact of steering strength on Qwen3-VL-4B.}
\label{tab:steering_strength}
\begin{tabular}{ccc}
\toprule
\(\boldsymbol{\lambda}\) 
& \textbf{Retain} \(\textcolor{red}{\boldsymbol{\uparrow}}\) 
& \textbf{Refusal Rollout Rate} \(\textcolor{red}{\boldsymbol{\uparrow}}\) \\
\midrule
0.2 & 0.54 & 0.01 \\
0.3 & 0.54 & 0.01 \\
1.0 & 0.54 & 0.01 \\
3.0 & \textbf{0.54} & \textbf{0.05} \\
\bottomrule
\end{tabular}
\end{table}

\begin{table}[t]
\centering
\caption{Hyperparameters Settings of Activation Steering}
\label{Training Configurations}
\begin{tabular}{lcccc} 
\toprule
\textbf{Model} & \textbf{Learning Rate} & \textbf{Batch Size} & \textbf{Epochs} & \textbf{$\lambda$} \\ 
\midrule
\textbf{Qwen3-VL-4B}  & 1e-4 & 4 & 3 & 3   \\
\textbf{Qwen3-VL-8B}  & 1e-4 & 4 & 3 & 0.2 \\
\textbf{LLaVA-1.5-7B} & 1e-4 & 4 & 3 & 0.8 \\
\bottomrule
\label{Hyperparameters Settings of Activation Steering}
\end{tabular}
\end{table}

\begin{table}[htbp]
\centering
\caption{Key hyperparameters for Refusal Boundary Optimization Stage}
\label{Refusal Boundary Optimization Training Configurations}
\begin{tabular}{lcccc} 
\toprule
\textbf{Model} & \textbf{Learning Rate} & \textbf{KL Coef} & \textbf{Actor Batch} & \textbf{Steps} \\ 
\midrule
\textbf{Qwen3-VL-4B}  & 1e-6 & 0.1 & 2 & 90  \\
\textbf{Qwen3-VL-8B}  & 1e-6 & 0.1 & 2 & 115 \\
\textbf{LLaVA-1.5-7B} & 1e-6 & 0.1 & 2 & 120 \\
\bottomrule
\end{tabular}
\end{table}
\subsection{Key hyperparameters for Refusal Boundary Optimization}\label{tab: ASRU hyperparameters}
The training hyperparameters for each model of ASRU are shown in Table~\ref{Refusal Boundary Optimization Training Configurations} and ~\ref{Hyperparameters Settings of Activation Steering}.
\section{Unlearning performance under different forgetting ratios}\label{appendix:unlearning with different ratios}
\begin{figure}[htbp]
    \centering

    \begin{subfigure}[t]{0.49\linewidth}
        \centering
        \includegraphics[width=\linewidth]{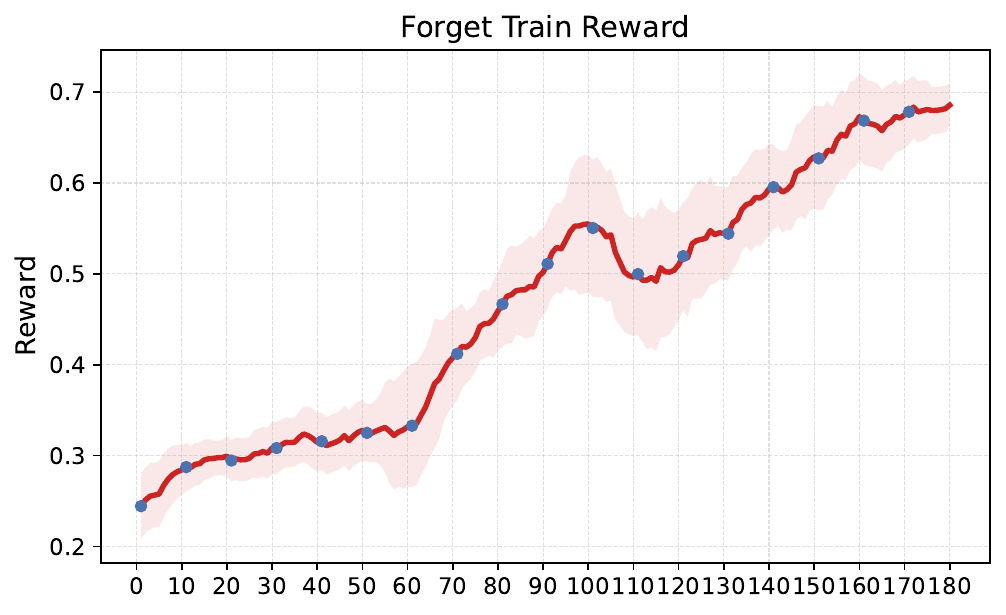}
        \caption{ Reward curve for 5\% ratios}
        \label{fig:reward1}
    \end{subfigure}
    \hfill
    \begin{subfigure}[t]{0.49\linewidth}
        \centering
        \includegraphics[width=\linewidth]{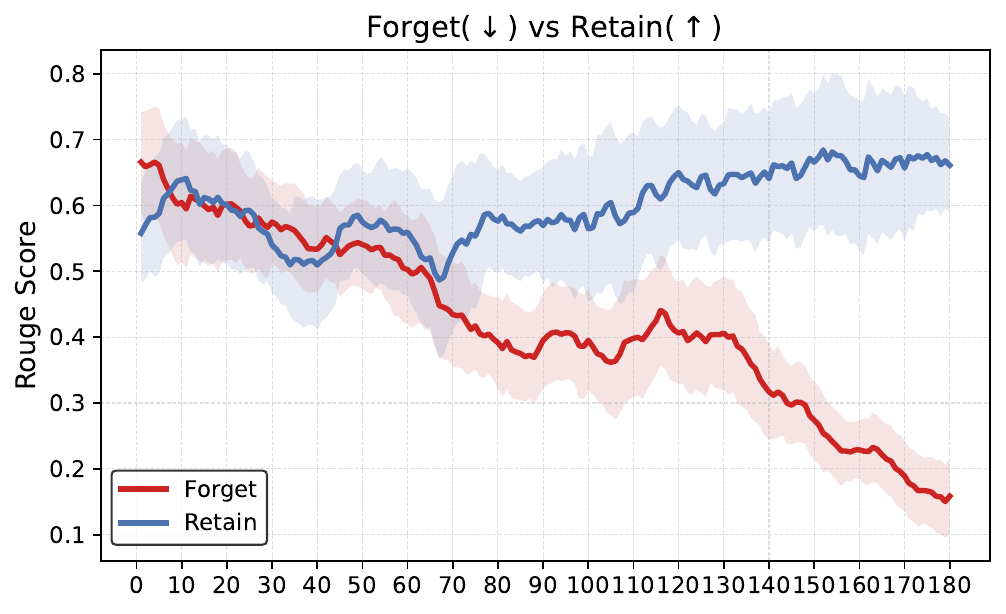}
        \caption{ROUGE scores for 5\% ratios}
        \label{fig:rouge1}
    \end{subfigure}

    \vskip\baselineskip 

    \begin{subfigure}[t]{0.49\linewidth}
        \centering
        \includegraphics[width=\linewidth]{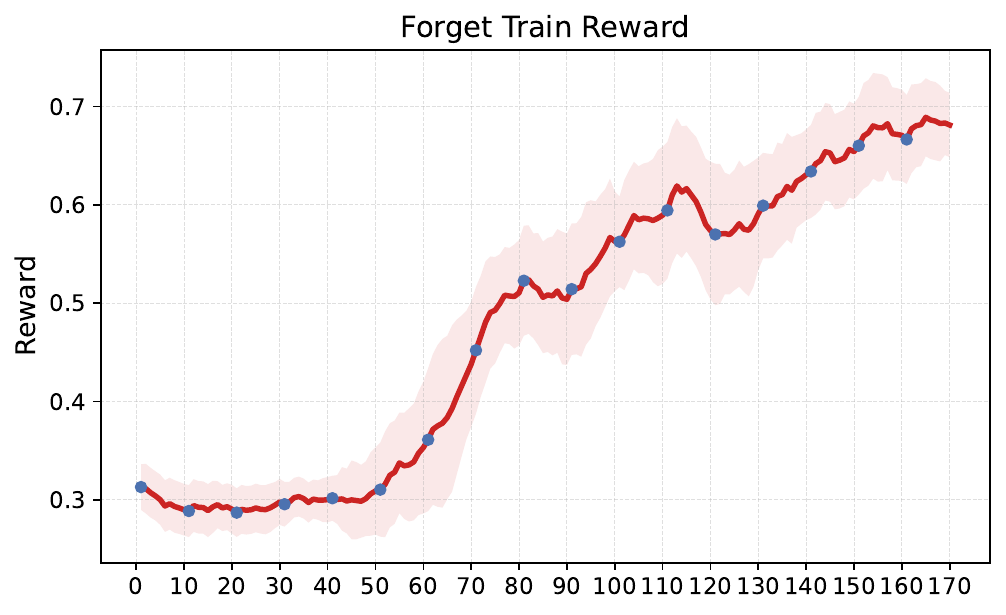}
        \caption{Reward curve for 10\% ratios}
        \label{fig:reward2}
    \end{subfigure}
    \hfill
    \begin{subfigure}[t]{0.49\linewidth}
        \centering
        \includegraphics[width=\linewidth]{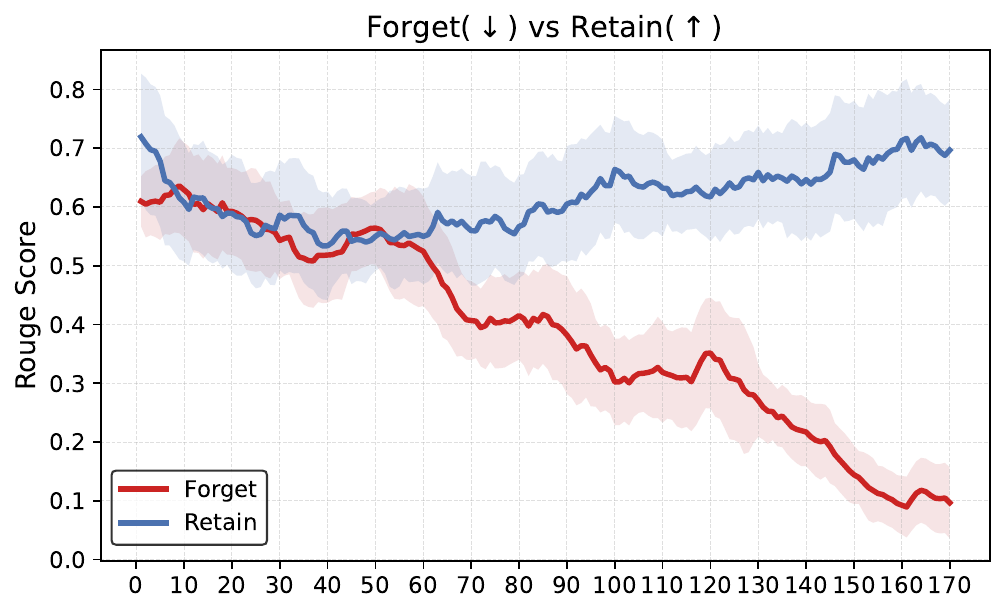}
        \caption{ROUGE scores for 10\% ratios}
        \label{fig:rouge2}
    \end{subfigure}

    \vskip\baselineskip 

    \begin{subfigure}[t]{0.49\linewidth}
        \centering
        \includegraphics[width=\linewidth]{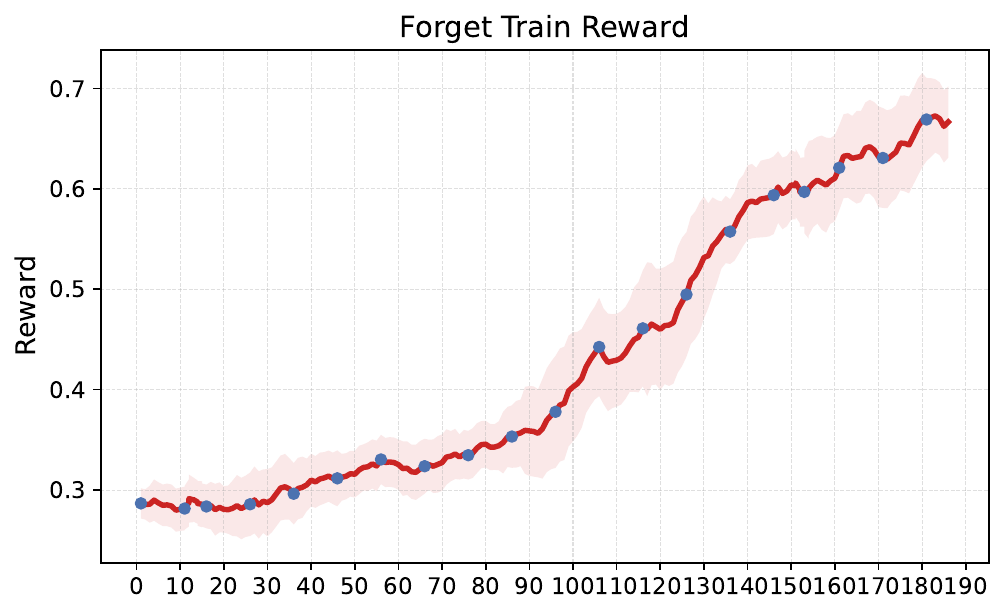}
        \caption{Reward curve for 15\% ratios}
        \label{fig:reward3}
    \end{subfigure}
    \hfill
    \begin{subfigure}[t]{0.49\linewidth}
        \centering
        \includegraphics[width=\linewidth]{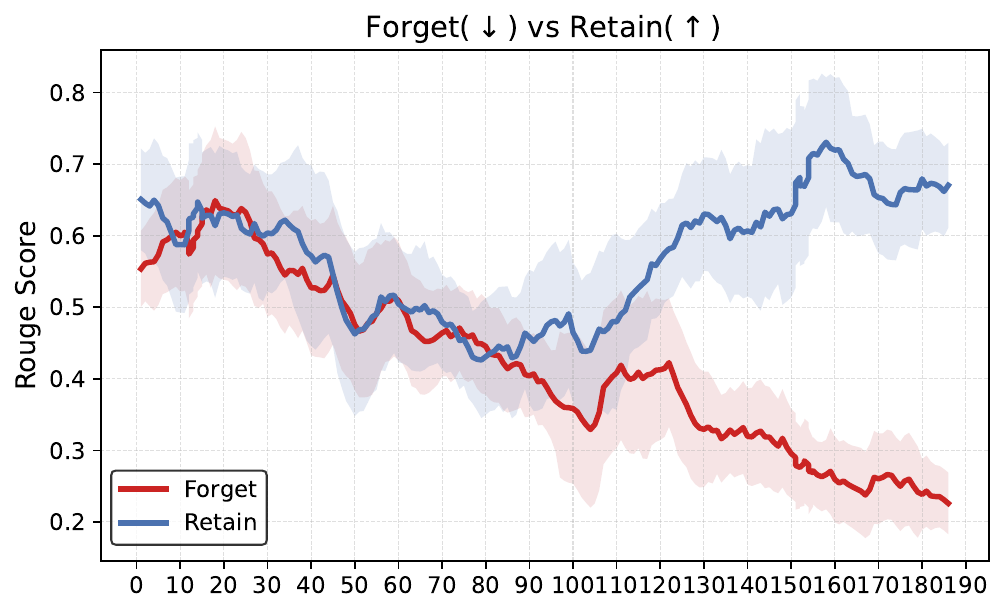}
        \caption{ROUGE scores for 15\% ratios}
        \label{fig:rouge3}
    \end{subfigure}

    \caption{The reward curves and changes in ROUGE scores for Qwen3-VL-8B under different forget ratio settings.}
    \label{different forget ratio settings.}
\end{figure}

As shown in table \ref{forget_ratio}, ASRU maintains high values in both Ret Acc (Retention Accuracy) and Real Acc (Real Accuracy). Notably, under the 15\% forget setting, ASRU exhibits strong recovery and retention capabilities, with both Ret Acc and Real Acc showing impressive performance. Despite forgetting target knowledge, ASRU outperforms most baseline methods in terms of accuracy on both the retain and real sets. This demonstrates its ability to effectively forget while maintaining or recovering model utility.Additionally, ASRU performs well on both Ret Rouge Score and Real Rouge Score. In the 15\% forget setting, ASRU achieves higher Ret Rouge Score and Real Rouge Score than other methods, reflecting its effective retention of non-target knowledge and high quality in generation tasks. Compared to other methods, ASRU maintains strong forgetting ability while also retaining competitive ROUGE scores, further proving its advantage in balancing forgetting and utility.Furthermore, as the forget ratio increases, ASRU demonstrates superior generation quality compared to the baseline methods.
Overall, ASRU achieves an excellent balance between forget and utility retention through the joint optimization of activation steering and GRPO, ensuring strong performance across different settings.
We also present the reward curves and changes in ROUGE scores for Qwen3-VL-8B under different forget ratio settings in Figures \ref{different forget ratio settings.}.

\section{Evaluating ASRU On CLEAR Datasets}\label{results: CLEAR Datasets}
To further validate the generalizability of ASRU, we conduct experiments on the CLEAR datasets.
As shown in Table~\ref{tab:clear_results}, ASRU also shows excellent performance on CLEAR. Specifically, it achieves gains of 4.79\% and 51.61\% in classification and generation performance, respectively, on the forget set, while maintaining utility on the real-world set, further validating the effectiveness and generalizability of our method.

\section{Additional Experiments}\label{Additional Experiments}
We also explored the effect of another training strategy, where each forget and retain sample is concatenated into a single prompt for training, using the same reward function as ASRU (encouraging the rejection of forget samples and the retention of retain samples). By placing \(D_f\) and \(D_r\) in the same context, the model simultaneously sees the two objectives:reject (forget) and accept (retain). As shown in Table \ref{tab:additional_experiments}, under the 15\% forget ratio setting, the ``mix'' strategy shows an improvement in forget quality compared to separate training, which aligns with intuition. By placing ``reject/accept" together in the same prompt, it resembles training a context-conditioned behavior switch, making rejection responses more contextually appropriate. The generation quality also improves significantly in certain forget ratios, though it lacks stability. However, the cost of ``mi'' is also clear: it tends to sacrifice performance on the retain set. At higher forgetting strengths, the joint context may introduce gradient interference, leading to a decline in retain quality. Therefore, ``mix'' is a promising but more ``aggressive forgetting/stronger rejection generation" strategy. Overall, separate mixed training (ASRU) is more inclined towards a ``robust retention utility" strategy.

\begin{table*}[t]
\centering
\caption{Unlearning performance on MLLMU-Bench (10\% forget and 15\% forget) on Qwen3-VL-8B-Instruct. 
\textcolor{red}{$\boldsymbol{\downarrow}$} indicates that lower values are preferred, while \textcolor{red}{$\boldsymbol{\uparrow}$} indicates that higher values are preferred.}
\resizebox{\linewidth}{!}{
\renewcommand\arraystretch{1.1}
\begin{tabular}{lccccccccccc}
\toprule
\multirow{2}{*}{\textbf{Method}} & \multirow{2}{*}{\textbf{samples}}  & \multicolumn{4}{c}{\textbf{Forget Quality} \textcolor{red}{$\boldsymbol{\downarrow}$ }}                                                         & \multicolumn{2}{c}{\multirow{2}{*}{\textbf{Generation Quality} \textcolor{red}{$\boldsymbol{\uparrow}$ }}} & \multicolumn{4}{c}{\textbf{Retain Quality} \textcolor{red}{$\boldsymbol{\uparrow}$ }}                                                                   \\ \cmidrule(lr){3-6} \cmidrule(lr){9-12}
\textbf{}            & \textbf{}         & \multicolumn{2}{c}{\textbf{Class.}} & \multicolumn{2}{c}{\textbf{Gen.}} & \multicolumn{2}{c}{}                                             & \multicolumn{2}{c}{\textbf{Class.}} & \multicolumn{2}{c}{\textbf{Gen.}} \\ \midrule
\textbf{}            & \textbf{$\tilde{D_r}$} & \textbf{Forget}              & \textbf{Test}             & \textbf{Forget}          & \textbf{Test}           & \textbf{CR}       & \textbf{Forgetfulness}       & \textbf{Retain}             & \textbf{Real}              & \textbf{Retain}          & \textbf{Real}           \\\midrule

\rowcolor{gray!10} \multicolumn{12}{c}{\textbf{Qwen3-VL-8B(10\%Forget)-VQA}}         \\ 

\textcolor{gray}{\textbf{Vanilla}}     &\textcolor{gray}{-}                 & \textcolor{gray}{45.71}                       & \textcolor{gray}{54.29}                      & \textcolor{gray}{0.569}                    &\textcolor{gray}{ 0.355}                   &\textcolor{gray}{ -}                                 & \textcolor{gray}{-}                            & \textcolor{gray}{52.99 }                      &\textcolor{gray}{ 78.59 }                     &\textcolor{gray}{ 0.570}                   &\textcolor{gray}{ 0.434 }                  \\
\textbf{GA}          & 0.00\%               & 46.12                       & 53.06                      & 0.569                    & 0.369                   & 0.01                              & 0.01                         & 53.08                       & 71.54                      & 0.556                    & 0.387                   \\
\textbf{GA\_diff}    & 100\%             & 46.12                       & 51.43                      & 0.523                    & 0.330                   & 0.05                              & 0.05                         & 49.78                      & 72.06                      & 0.541                    & 0.386                   \\
\textbf{NPO}         & 0.00\%               & 47.75                       & 54.69                      & 0.461                    & 0.331                   & 0.01                              & 0.04                        & 51.07                       & 72.54                      & 0.460                    & 0.385                   \\
\textbf{KL\_Min}     & 100\%             & 43.27                       & 53.06                      & 0.574                    & 0.326                  & 0.07                              & 0.09                         & 49.95                       & 71.48                      & 0.552                    & 0.374                   \\
\textbf{MMunlearner} & 100\%          & 56.67                       & 60.83                     & 0.540                    & 0.315                   & 0.30                              & 0.34                         & 52.96                       & 73.11                      & 0.526                    &0.376                    \\
\rowcolor{blue!10} \textbf{ASRU(ours)}  & 28.80\%           & \textbf{35.92}              & \textbf{48.97}             & \textbf{0.271}           & \textbf{0.257}          & \textbf{4.54}                     & \textbf{4.41}                & \textbf{53.93}              & \textbf{73.24}             & \textbf{0.559}           & \textbf{0.389}          \\ \midrule
\rowcolor{gray!10} \multicolumn{12}{c}{\textbf{Qwen3-VL-8B(15\%Forget)-VQA}}                                                                                                                                                                                                                                                                                    \\
                
\textcolor{gray}{\textbf{Vanilla}}     &\textcolor{gray}{ -}                 & \textcolor{gray}{45.33}                       & \textcolor{gray}{54.13}                      & \textcolor{gray}{0.555}                    &\textcolor{gray}{ 0.322}                   &\textcolor{gray}{ -}                                 &\textcolor{gray}{ - }                           & \textcolor{gray}{49.10}                       &\textcolor{gray}{ 78.59}                      &\textcolor{gray}{ 0.572}                     & \textcolor{gray}{0.434 }                   \\
\textbf{GA}          & 0.00\%               & 45.06                       & 52.80                      & 0.549                    & 0.338                   & 0.17                              & 0.19                         & 47.97                       & 74.15                      & 0.541                    & 0.364                   \\
\textbf{GA\_diff}    & 100\%             & 44.80                       & 53.60                      & 0.549                    & 0.335                   & 0.17                              & 0.17                         & 46.32                        & 73.50                      & 0.534                    & 0.379                   \\
\textbf{NPO}         & 0.00\%               & 45.60                       & 48.00                      & 0.406                    & 0.270                   & 0.03                              & 0.05                         & 47.41                       & 76.76                      & 0.414                    & 0.355                   \\
\textbf{KL\_Min}     & 100\%             & 41.07                       & 49.07                      & 0.537                    & 0.321                   & 0.26                              & 0.27                         & 44.72                       & 76.11                       & 0.523                    & 0.371                   \\
\textbf{MMunlearner} & 100\%           & 49.33                       & 59.20                      & 0.551                    & 0.336                   & 0.093                              & 0.12                         & 51.42                       & 76.76                      & 0.538                    & 0.379                   \\
\rowcolor{blue!10} \textbf{ASRU(ours)}  & 30.55\%           & \textbf{30.67}              & \textbf{41.33}             & \textbf{0.339}           & \textbf{0.253}          & \textbf{2.77}                     & \textbf{2.77}                & \textbf{57.97}              & \textbf{77.02}             & \textbf{0.548}           & \textbf{0.380}         \\ \bottomrule
\label{forget_ratio}
\end{tabular}}
\end{table*}


\section{Case Study}\label{app:case study}
\begin{figure}[t]
    \centering
        \includegraphics[width=\linewidth]{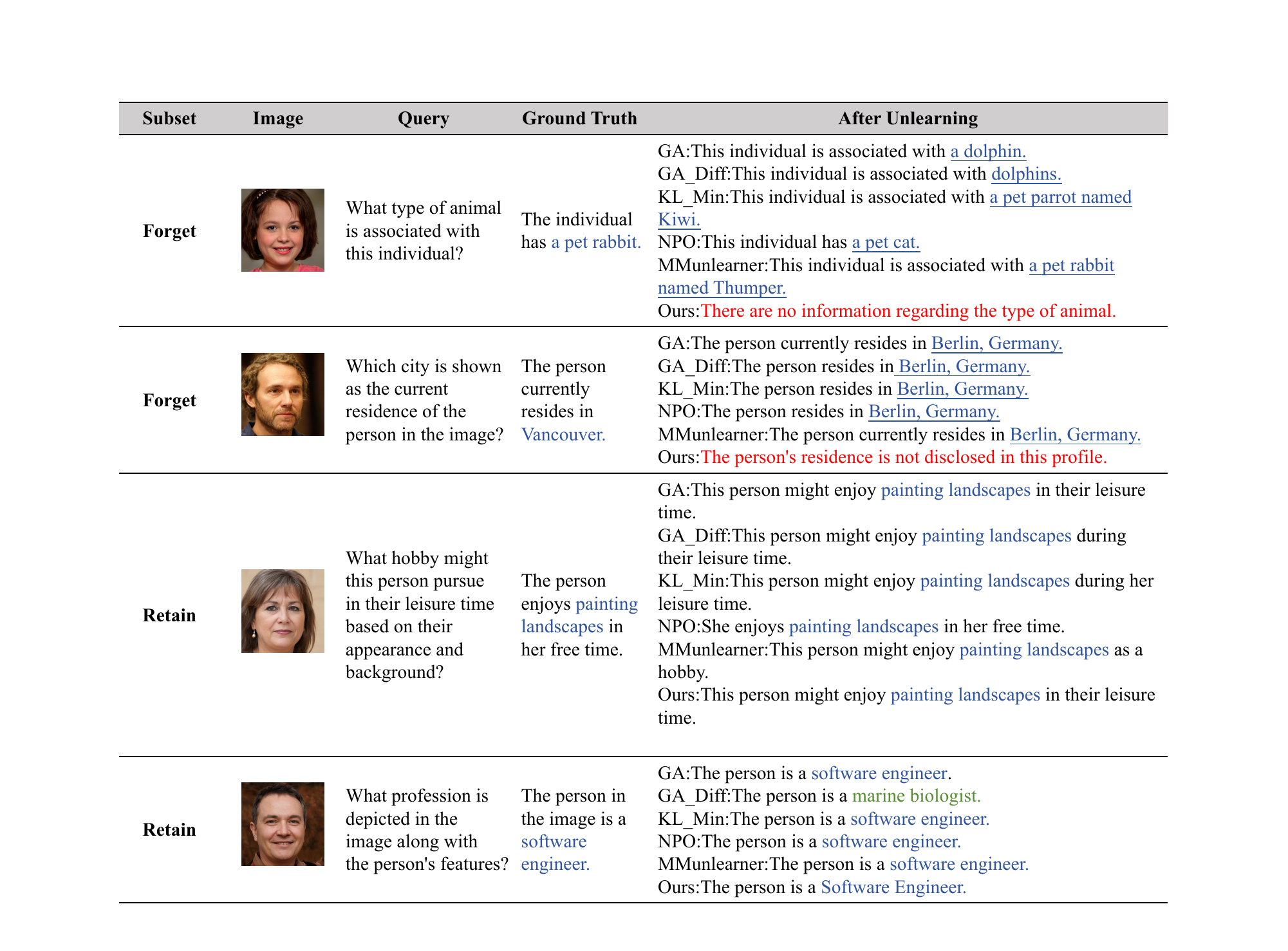}
    \caption{Case Study on Forget Set and Retain set before and after unlearning}
    \label{case study}
\end{figure}

To provide a more intuitive understanding of the effects of different unlearning methods, we present case studies on the forget and retain sets in the figure. These examples illustrate the performance of various methods before and after unlearning. As shown in Figure \ref{case study}, most unlearning methods expose forget information during the unlearningf process, resulting in suboptimal forget effects. However, our method successfully removes the target information and generates contextually appropriate rejection responses. In terms of maintaining model utility, our method preserves the originally correct responses on the retain set, demonstrating its superior ability to effectively perform unlearning while maintaining retained knowledge.

\end{document}